\documentclass[11pt]{article}

\usepackage[preprint]{acl}

\usepackage{times}
\usepackage{latexsym}
\usepackage{natbib}  % DO NOT CHANGE THIS AND DO NOT ADD ANY OPTIONS TO IT
\usepackage{caption} % DO NOT CHANGE THIS AND DO NOT ADD ANY OPTIONS TO IT
\usepackage{multirow}
\usepackage{algorithm}
\usepackage{algorithmic}
\usepackage{amsmath}
\usepackage{newfloat}
\usepackage{listings}
\usepackage{amssymb}

\DeclareCaptionStyle{ruled}{labelfont=normalfont,labelsep=colon,strut=off} % DO NOT CHANGE THIS
\floatstyle{ruled}
\newfloat{listing}{tb}{lst}{}
\floatname{listing}{Listing}

\usepackage{booktabs}
\usepackage[T1]{fontenc}
\usepackage[utf8]{inputenc}

\usepackage{microtype}

\usepackage{inconsolata}

\usepackage{graphicx}

\title{Hidden Cartographer: Hidden State Guided Data Selection for Dual Budgeted Instruction Tuning}

\author{Zixuan Qin\textsuperscript{1}, 
   Qi Qin\textsuperscript{1}, 
    Erbo Li\textsuperscript{1},
    Ting Wei\textsuperscript{1},
    Yifan Sun\textsuperscript{1*},
    Zhaoxin Fan\textsuperscript{2*}
     \\
\textsuperscript{1}Center for Applied Statistics, School of Statistics, Renmin University of China \\
\textsuperscript{2}Beijing Advanced Innovation Center for Future Blockchain and Privacy Computing }

\begin{document}
\maketitle
\begin{abstract}
% Deploying private LLMs in specialized domains such as medicine, law, and finance is constrained by both data privacy and resource budgets: private-domain data cannot be freely exposed for large-scale expert annotation, while full-dataset fine-tuning is computationally expensive. Existing data selection methods are difficult to apply in this setting: label-dependent methods become unreliable when ground-truth labels are replaced by noisy pseudo-labels, while traditional active learning methods optimize generic diversity or uncertainty signals that are not aligned with LLM training dynamics. We propose Hidden Cartographer, a hidden-state-guided data selection framework for privacy-constrained private-domain LLM adaptation. The key insight is that LLM hidden states provide a label-free proxy for estimating confidence and variability before exhaustive annotation. HC first trains a lightweight proxy from a small expert-labeled seed set to expand a high-value candidate pool from unlabeled private data. It then computes exact confidence-variability scores only within this pool and prunes it into a compact final subset, reducing fine-tuning cost. Across medical, legal, and financial instruction-tuning scenarios, HC consistently achieves the best or near-best performance under the same budget, while reducing expert annotation cost by 80\% and final training cost by 90\%.

Given the high cost of expert annotation and limited fine-tuning resources, effective data selection from large unlabeled pools under strict budgets has become a crucial step in adapting LLMs to specialized domains. Existing label-dependent methods risk distorted selection when using pseudo-labels, while traditional active learning methods do not explicitly characterize instruction-tuning dynamics. To address these limitations, we propose Hidden Cartographer (HC), a two-stage framework that uses hidden states to guide annotation and exact training dynamics to select the final training subset. The first stage is Progressive Candidate Expansion, which utilizes the proxy to iteratively select top unlabeled candidates for expert annotation, merging them to recompute training dynamics and update the proxy until the annotation budget is exhausted; the second stage is Exact Cartography Pruning, which ranks these annotated candidates by their exact dynamics scores to isolate a compact subset under the final training budget. Experiments across the medical, legal, and financial domains demonstrate that HC achieves the highest average accuracy among all budgeted baselines, requiring annotation for only 20--30\% of the original pool and fine-tuning on just 10--15\%. 

% Our code is available at \url{https://anonymous.4open.science/r/Hidden-Cartographer}.

\end{abstract}

\begingroup
\renewcommand\thefootnote{} 
\begin{NoHyper} 
\footnotetext{*Project leaders and Corresponding authors. Emails: sunyifan@ruc.edu.cn, zhaoxinf@buaa.edu.cn}
\end{NoHyper} 
\endgroup

\section{Introduction}

Large language models perform well across a range of NLP tasks~\cite{yu2026cascadehierarchicalrecoverabilitycontrol,li2026maskfreeprivacyextractionrewriting}. However, achieving optimal performance in vertical domains such as medicine, law, and finance often requires instruction tuning on domain data, particularly for smaller scale models~\cite{ke-etal-2025-demystifying,slack-al-moubayed-2025-early,li2024quantityqualityboostingllm}. In practice, this process is subject to two budget constraints: domain data are abundant and unlabeled, and obtaining high-quality annotations requires costly expert knowledge; fine-tuning on all available data consumes substantial computational resources~\cite{xia-etal-2025-selection}. Data selection provides an effective way to overcome these constraints: it can prioritize unlabeled samples to guide expert annotation and select the most valuable annotated samples for final fine-tuning. We therefore study how to construct an efficient training subset from a large unlabeled domain pool under joint constraints on annotation cost and computational resources.

\begin{figure}[t]
\centering
\includegraphics[width=\columnwidth]{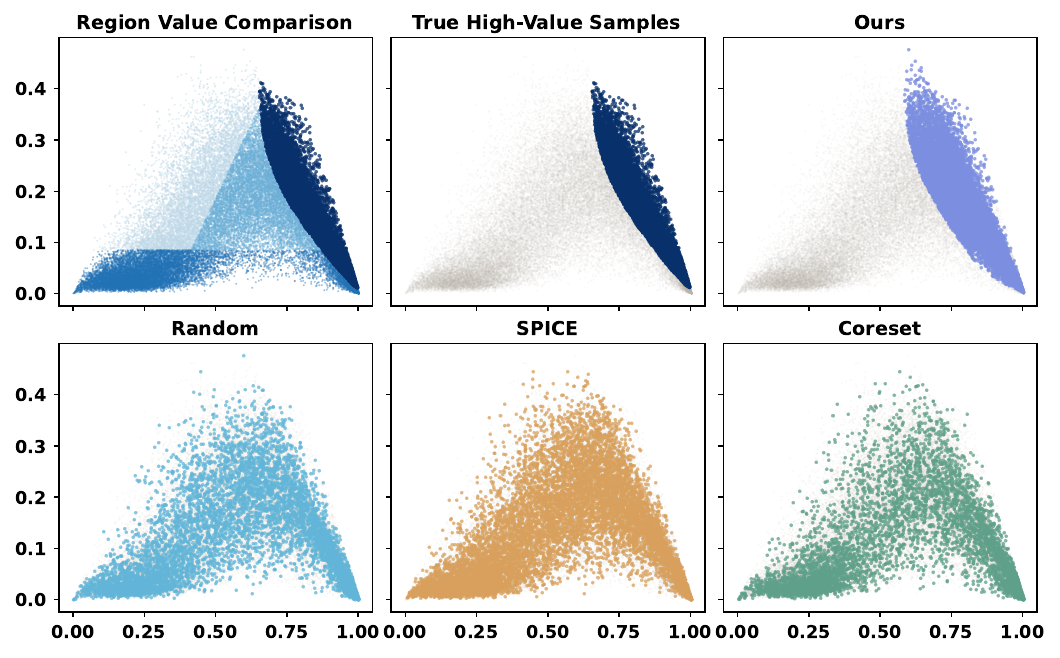}
\caption{Training-data value and selection distributions in the medical domain. All panels use the same Dataset Cartography space, with confidence on the x-axis and variability on the y-axis. Top left: Four regional subsets are constructed by sampling 15\% of the full training pool from each region. Darker blue indicates better performance on medical benchmarks after fine-tuning on the corresponding subset. Top middle: The best-performing region among the four, identified as the region containing high-value samples. Other panels: Distributions of samples selected by HC, Random, SPICE, and Coreset at the same 15\% selection ratio.}
\label{motivation}
\vspace{-4mm}
\end{figure}

Existing data selection methods typically rely on labels to compute losses and gradients or assess samples' influence on model utility~\cite{xia2024lessselectinginfluentialdata,deb2025fishersft,lin2025leaditerativedataselection,chang2026spice,li2024superfilteringweaktostrongdatafiltering,yang2024smalltolarge}. However, this paradigm requires annotating the entire unlabeled pool in advance, making it difficult to satisfy practical budget constraints. Although pseudo-labeling allows these methods to be directly applied to unlabeled data acquisition, model-generated responses often contain domain-specific factual errors, which can introduce additional noise and distort the resulting selection signals. Meanwhile, though traditional active learning methods~\cite{sener2018active,Ash2020Deep,ash2021gonefishingneuralactive,kung2023activeinstructiontuningimproving} select unlabeled candidates using heuristic criteria such as diversity and uncertainty, these criteria characterize only properties of the data distribution and do not explicitly model candidates' dynamic learning behavior during instruction tuning. Consequently, the key challenge is to accurately estimate a selection signal closely linked to downstream model utility before labels are available.

To address this challenge, we adopt a training dynamics perspective to establish a selection signal that reflects the target model's learning behavior. Building on Data Cartography~\cite{swayamdipta-etal-2020-dataset}, we monitor token-length-normalized response probabilities across training checkpoints to derive sample confidence and variability for instruction tuning (detailed in Section~\ref{Preliminary}). As illustrated in Figure~\ref{motivation}, these metrics effectively isolate a high-utility data region that existing selection methods fail to target. While exact training dynamics provide an ideal selection criterion, computing them inherently requires ground-truth labels. To estimate this signal prior to annotation, we leverage prompt-only hidden representations extracted directly from the frozen target model. Prior research indicates that such hidden states intrinsically encode model-perceived difficulty~\cite{llmalreadyknows} and semantic uncertainty~\cite{ka2l}. We hypothesize that these representations can reliably predict the target model's training dynamics using only unlabeled inputs.

Based on this insight, we propose Hidden Cartographer (HC), a two-stage framework for data selection under both annotation and training constraints. The first stage, Progressive Candidate Expansion, addresses the annotation constraint. HC starts from a small expert-labeled seed set, extracting hidden-state features to train a proxy predicting confidence and variability. Due to limited initial labels, HC avoids a one-shot final data selection. Instead, it iteratively expands a candidate pool via active learning: the proxy scores unlabeled samples, experts annotate the top-K selections, and this newly labeled data refines the proxy for the next round. This process improves the recall of high-value samples within a limited annotation budget.

The second stage, Exact Cartography Pruning, addresses the training cost constraint. Once the candidate pool is annotated, HC computes the exact confidence and variability within this smaller pool and applies a scoring function. HC then selects the final subset for instruction tuning from the candidate pool. This design connects the two stages: hidden states are used to cheaply recover a high-quality candidate pool from the unlabeled data, and exact cartography is used to precisely prune this pool into a much smaller training set. In this way, HC avoids relying on noisy pseudo-labels for full-pool selection while also avoiding the cost of annotating and training on the entire dataset.

Our contributions are summarized as follows:
\begin{itemize}
    \item We formulate domain LLM data selection as a dual-constrained problem, where both expert annotation and LLM training computation are limited. We further identify LLM hidden states as a label-free signal for estimating sample training dynamics without relying on noisy pseudo-labels or full-dataset annotation.

    \item We propose Hidden Cartographer, a two-stage framework with Progressive Candidate Expansion and Exact Cartography Pruning. The first stage uses a hidden-state proxy to recover high-value samples under limited annotation, while the second computes exact confidence-variability scores to select a training subset.

    \item Experiments across medicine, law, and finance show that HC outperforms random selection and strong data selection baselines under the same budget. HC reduces expert annotation cost by 70--80\% by annotating only a candidate pool, and training cost by 85--90\% by fine-tuning on only 10--15\% of the data.
\end{itemize}

\section{Related Work}
\subsection{Data Selection for Instruction Tuning}
Data selection for instruction tuning uses signals such as quality, diversity, and model behavior~\cite{albalak2024surveydataselectionlanguage}. Representative approaches include feature-based quality estimation~\cite{cao2024instructionmininginstructiondata}, diversity-based selection with DPP~\cite{zhang2023dppbasedclientselectionfederated}, task-aware distribution matching in TSDS~\cite{liu2024tsds}, instruction-following difficulty (IFD)~\cite{li-etal-2024-quantity}, information gain in FisherSFT~\cite{deb2025fishersft}, gradient-based influence in LESS~\cite{xia2024lessselectinginfluentialdata}, dynamic uncertainty in LEAD~\cite{lin2025leaditerativedataselection}, uncertainty-aware self-reflection in SelectIT~\cite{liu2025selectitselectiveinstructiontuning}, gradient-conflict-aware selection in SPICE~\cite{chang2026spice}, weak-model IFD filtering with Superfilter~\cite{li2024superfilteringweaktostrongdatafiltering}, and small-model loss-trajectory clustering with S2L~\cite{yang2024smalltolarge}. These methods primarily select from existing instruction--response datasets, with many relying on response-conditioned signals, whereas HC guides expert annotation from an unlabeled pool and prunes the resulting candidates under separate annotation and final-training budgets.

\subsection{Active Learning for LLMs}
Active learning reduces annotation cost by selecting unlabeled samples using uncertainty, diversity, or both. Representative methods include uncertainty sampling, Coreset~\cite{sener2018active}, BADGE~\cite{Ash2020Deep}, and AIT~\cite{kung2023activeinstructiontuningimproving}. These methods are effective when the acquisition objective is aligned with the downstream learning signal. However, for LLM instruction tuning, gradient-based acquisition is costly, and diversity or uncertainty may not reflect whether a sample is useful for improving the model. In particular, covering the representation space does not necessarily recover samples with favorable training dynamics. In contrast, our work uses LLM hidden states as a label-free proxy for dynamics-based sample value, enabling candidate expansion under limited expert annotation.

\section{Preliminary: Training Dynamics Score}
\label{Preliminary}

We use training dynamics to evaluate each instruction-tuning sample. \citet{swayamdipta-etal-2020-dataset} proposed Dataset Cartography for classification tasks, tracking the probability assigned to the true label across training epochs. Since instruction tuning generates token sequences rather than single labels, we extend these metrics to the generative setting. Given a sample $(x_i, y_i^\ast)$, where $x_i$ denotes the instruction and input, and $y_i^\ast = (y_{i,1}^\ast,\ldots,y_{i,T_i}^\ast)$ is the target response, we track the probability of generating this response across $E$ checkpoints $\theta^{(1)},\ldots,\theta^{(E)}$. For checkpoint $\theta^{(e)}$, we define the token-normalized output probability as:
\begin{equation}
p_i^{(e)}
=
\exp \left(
\frac{1}{T_i}
\sum_{t=1}^{T_i}
\log p_{\theta^{(e)}}(y_{i,t}^\ast \mid x_i, y_{i,<t}^\ast)
\right).
\end{equation}
Token normalization mitigates length bias across responses. In practice, we compute probabilities only over the output tokens, masking out the instruction and input. Based on per-checkpoint probabilities, we define confidence and variability as:
\begin{equation}
C_i = \frac{1}{E} \sum_{e=1}^{E} p_i^{(e)}, 
\hspace{1mm}
V_i =
\sqrt{
\frac{1}{E}
\sum_{e=1}^{E}
\left(p_i^{(e)} - C_i\right)^2
}.
\end{equation}

Here, confidence reflects how well the model learns the sample, while variability captures the instability of its predictions during training. High-value samples should be learnable but non-trivial, providing informative signals rather than noise.

To combine confidence and variability on comparable scales, we apply min--max normalization independently within the current selection pool $\mathcal{P}$, denoted by $(\tilde{C}, \tilde{V}) = \mathrm{MinMaxScaler}(C, V)$. In Phase 1, we normalize the proxy predictions over the remaining unlabeled pool at each round; in Phase 2, we normalize the exact statistics over the annotated candidate pool.

We introduce an extreme penalty based on these normalized metrics and define the final training dynamics score:
\begin{equation}
P_{\mathrm{ext}}^{(i)}
=
\left|\tilde{C}_i - C_{\mathrm{med}}\right|
+
\left|\tilde{V}_i - V_{\mathrm{med}}\right|,
\end{equation}

\vspace{-5mm}

\begin{equation}
\mathrm{Score}_i
=
\alpha \tilde{C}_i
+
\beta \tilde{V}_i
-
\gamma P_{\mathrm{ext}}^{(i)}.
\end{equation}
where $C_{\mathrm{med}}$ and $V_{\mathrm{med}}$ are the medians of $\tilde{C}$ and $\tilde{V}$ over the current selection pool. This term penalizes samples with abnormal training dynamics by measuring deviation from $C_{\mathrm{med}}$ and $V_{\mathrm{med}}$.

% \section{Limitation of Existing Methods}
% \input{3-challenge}

\section{Methodology: Hidden Cartographer}
\begin{figure*}[htbp]
\centering
\includegraphics[width=1.0\textwidth]{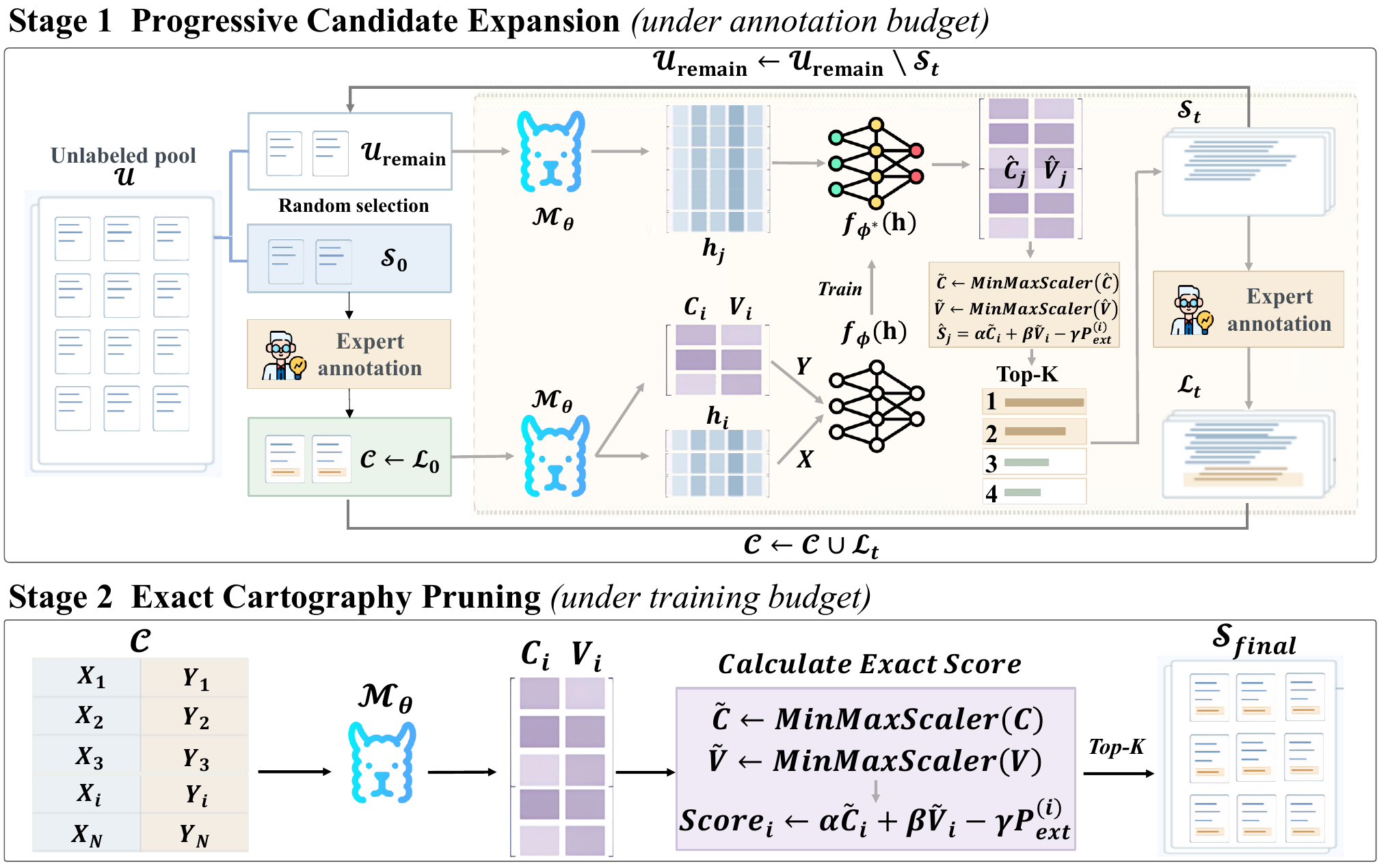}
\caption{Overview of the Hidden Cartographer framework.}
\label{workflow}
\vspace{-4mm}
\end{figure*}

In this section, we present HC, a two-stage framework for data selection under annotation and fine-tuning budgets. HC progressively selects samples from a large unlabeled pool for expert annotation, then prunes the annotated candidate pool into a subset for final fine-tuning. Detailed implementation for both stages is provided in Appendix~\ref{Algorithm Implementation}.

\subsection{Problem Formulation}
We are given a large unlabeled pool $\mathcal{U}=\{x_1,\ldots,x_M\}$ of domain-specific inputs and an expert annotation oracle $\mathcal{O}$ that provides the response $y_i^\ast$ for a queried input $x_i$. We consider two budgets, both measured in numbers of examples: $B_{\mathrm{ann}}$ for expert annotation and $B_{\mathrm{train}}$ for final fine-tuning, with $B_{\mathrm{train}}\le B_{\mathrm{ann}}\ll M$. Seed annotations are included in $B_{\mathrm{ann}}$. Let $\pi$ denote a selection strategy comprising adaptive annotation and final pruning. Each annotation decision selects a previously unlabeled input using the available inputs and acquired annotations. Queries to $\mathcal{O}$ produce an annotated candidate pool $\mathcal{C}_\pi$ of input--gold-response pairs, from which the strategy selects a final training subset $\mathcal{S}_\pi\subseteq\mathcal{C}_\pi$. Gold responses of unqueried inputs are unavailable to the strategy. Let $\theta^\ast(\mathcal{S}_\pi)$ denote the target LLM parameters obtained by applying a fixed fine-tuning procedure to $\mathcal{S}_\pi$. Our objective is to find an admissible strategy that minimizes expected loss on the target domain:
\begin{equation}
\begin{aligned}
\pi^\ast \in {}&
\mathop{\arg\min}_{\pi}
\mathbb{E}_{\pi}
\left[
R\!\left(\theta^\ast(\mathcal{S}_\pi)\right)
\right] \\
\text{s.t.}\quad
&|\mathcal{C}_\pi|\le B_{\mathrm{ann}},
|\mathcal{S}_\pi|\le B_{\mathrm{train}},\mathcal{S}_\pi \subseteq \mathcal{C}_\pi
\end{aligned}
\end{equation}
where $R(\theta)=\mathbb{E}_{(x,y^\ast)\sim\mathcal{D}_{\mathrm{target}}}[\ell(\mathcal{M}_{\theta}(x),y^\ast)]$, $\ell$ is the autoregressive cross-entropy loss, and the outer expectation accounts for any randomness in acquisition and pruning. The size and subset constraints apply to every run. Direct optimization is impractical because evaluating selection strategies requires repeated annotation and fine-tuning.

\subsection{HC Algorithm}

HC separates the process into two stages: Progressive Candidate Expansion, which uses limited annotation and hidden-state prediction to build a high-value candidate pool $\mathcal{C}$, and Exact Cartography Pruning, which applies exact training dynamics inside this candidate pool to select the final subset. The overall workflow is shown in Figure~\ref{workflow}.

The first phase, Progressive Candidate Expansion, aims to reduce the annotation space by recovering a candidate pool with high coverage of valuable samples. The process begins with seed initialization, where HC uniformly samples a small subset $\mathcal{S}_0$ from the unlabeled pool $\mathcal{U}$ for expert annotation. This produces the initial labeled set $\mathcal{L}_0$ and initializes the candidate pool $\mathcal{C}$, leaving the remaining samples in $\mathcal{U}_{\mathrm{remain}}$. Because only samples in $\mathcal{C}$ have target outputs, exact confidence and variability are initially computed exclusively within this small pool. Next, during proxy learning, HC tracks the output probability across training checkpoints for each annotated sample $(x_i,y_i) \in \mathcal{C}$ to compute its confidence $C_i$ and variability $V_i$. It simultaneously extracts the prompt-level hidden state $\mathbf{h}_i$ from the target LLM $\mathcal{M}_\theta$. The resulting tuples $(\mathbf{h}_i,C_i,V_i)$ are used to train a lightweight proxy $f_\phi$ that maps hidden states to training dynamics. Finally, HC applies this proxy to actively expand the pool. For each unlabeled sample $x_j \in \mathcal{U}_{\mathrm{remain}}$, it extracts the prompt-only hidden state $\mathbf{h}_j$, predicts the dynamics $(\hat{C}_j,\hat{V}_j)$, and computes the corresponding selection score. The top-scoring batch is annotated by experts, merged into $\mathcal{C}$, and removed from $\mathcal{U}_{\mathrm{remain}}$. Repeating this cycle iteratively refines the proxy and expands the candidate pool toward the high-value region without requiring full annotation of the dataset.

The second phase, Exact Cartography Pruning, focuses on reducing the final training cost. Although the candidate pool $\mathcal{C}$ is fully annotated after the first phase, it remains larger than the final fine-tuning budget. Furthermore, the initial scores are merely proxy estimates suited for candidate discovery rather than final selection. HC therefore recomputes the exact confidence and variability for every candidate sample using the target LLM and the newly available outputs. Because $\mathcal{C}$ is significantly smaller than the original pool, this exact cartography can be performed at a manageable computational cost. HC then ranks all candidate samples by their exact training dynamics score, favoring learnable and informative samples while penalizing overly easy, unstable, or noisy outliers. The top $N_{\mathrm{final}}$ samples are selected to form the final instruction-tuning set $\mathcal{S}_{\mathrm{final}}$.

These two stages serve complementary roles. The first phase uses hidden states to allocate the limited expert annotation budget efficiently, distilling a large unlabeled pool into a dense candidate pool. The second phase applies exact cartography to prune this pool into a compact training set. Consequently, HC combines the efficiency of proxy prediction for candidate discovery with the precision of exact dynamics for final selection.

\section{Theoretical Analysis}

In this section, we establish the theoretical properties of HC's selection. We first characterize score stability under proxy prediction errors, then derive sufficient conditions for progressive candidate coverage, and finally bound the reference-score regret of the final subset. Detailed assumptions and proofs are provided in Appendix~\ref{app:theory}.

\paragraph{Score Stability.}
On a nonempty finite pool $\mathcal{P}$, let $q_i$ and $\hat q_i$ be HC scores computed from exact and predicted raw dynamics, respectively. Each score vector uses its own pool-wide min--max normalization and its own confidence and variability medians. For $Z\in\{C,V\}$, define the raw range $r_Z=\max_{i\in\mathcal{P}}Z_i-\min_{i\in\mathcal{P}}Z_i$ and error $\varepsilon_Z=\max_{i\in\mathcal{P}}|\hat Z_i-Z_i|$.

\noindent\textbf{Theorem 1.}
\textit{For $\alpha,\beta,\gamma\ge0$, let
$\delta_Z=\min\{1,2\varepsilon_Z/r_Z\}$ if $r_Z>0$, and $\delta_Z=1$ otherwise. Then}
\begin{equation}
\begin{aligned}
\max_{i\in\mathcal{P}}|\hat q_i-q_i|&\le\eta,\\
\eta&=(\alpha+2\gamma)\delta_C
       +(\beta+2\gamma)\delta_V.
\end{aligned}
\label{eq:hc-score-stability}
\end{equation}
Both medians can move under prediction error. Thus, if the two normalized errors are at most $\delta$, the bound is $(\alpha+\beta+4\gamma)\delta$. Small raw errors do not necessarily guarantee small score errors when the underlying raw range is small.

\paragraph{Candidate Coverage.}
For notational convenience, we identify datasets with their example-index sets. Fix an exact HC reference score $Q_i$ on the original pool $\mathcal{U}$ under one specified trajectory; this reference is used exclusively for analysis. Let $G^\ast=\operatorname{TopK}_{\mathcal{U}}(Q)$, where $K=N_{\mathrm{final}}$, $1\le K\le\min\{B_{\mathrm{train}},|\mathcal{U}|\}$, and fix a common tie-breaking order. For a given seed $\mathcal{C}_0$, write $\mathcal{P}_0=\mathcal{U}\setminus\mathcal{C}_0$ and $m=|G^\ast\cap\mathcal{P}_0|$.

For the score-guided variant in Algorithm 1, round $t$ acquires the top $b_t$ predicted scores from the current remaining pool $\mathcal{P}_t$. Let $q_i^t$ be the exact comparator using that round's trajectory and normalization pool. Define
\begin{equation}
d_t=\max_{i\in\mathcal{P}_t}|q_i^t-Q_i|,
\quad
E_{\mathrm{acq}}=\max_t(\eta_t+d_t),
\end{equation}
where $\eta_t$ is the score-error bound from Theorem 1. The drift $d_t$ explicitly accounts for both trajectory updates, and median statistics.

\noindent\textbf{Theorem 2.}
\textit{Suppose acquisition uses distinct Top-$b_t$ queries, with $b=\sum_t b_t$ and $|\mathcal{C}_0|+b\le B_{\mathrm{ann}}$. For $1\le m\le b<|\mathcal{P}_0|$, the final candidate pool $\mathcal{C}$ contains $G^\ast$ if}
\begin{equation}
Q^{\mathcal{P}_0}_{(m)}
-
Q^{\mathcal{P}_0}_{(b+1)}
>
2E_{\mathrm{acq}},
\label{eq:hc-coverage-condition}
\end{equation}
\textit{where the order statistics are computed over the fixed scores $Q$ restricted to $\mathcal{P}_0$.}

Coverage trivially holds if $m=0$ or all remaining examples are annotated. For a fixed reference and error bound, oversampling permits separation between ranks $m$ and $b+1$, rather than requiring exact recovery at each round's boundary.

\paragraph{Final Score Regret.}
Let $q_i^{F}$ be the score recomputed on the final annotated pool $\mathcal{C}$ using the final cartography trajectory, and let $\mathcal{S}_{F}=\operatorname{TopK}_{\mathcal{C}}(q^{F})$. Define $D_F=\max_{i\in\mathcal{C}}|q_i^{F}-Q_i|$, $m_{\mathrm{miss}}=|G^\ast\setminus\mathcal{C}|$, and $W=\max_{\mathcal{U}}Q-\min_{\mathcal{U}}Q$.

\noindent\textbf{Theorem 3.}
\textit{If $|\mathcal{C}|\ge K$, then}
\begin{equation}
0\le
\sum_{i\in G^\ast}Q_i-\sum_{i\in\mathcal{S}_F}Q_i
\le
m_{\mathrm{miss}}W+2KD_F.
\label{eq:hc-two-stage-regret}
\end{equation}
\textit{For $K<|\mathcal{U}|$, if $G^\ast\subseteq\mathcal{C}$ and
$Q_{(K)}-Q_{(K+1)}>2D_F$, then $\mathcal{S}_F=G^\ast$.}

The term $m_{\mathrm{miss}}W$ bounds reference-score loss from omitted reference samples, while $2KD_F$ bounds effect of score changes caused by the training trajectory and pool statistics. Exact pruning uses exact dynamics within the candidate pool and recovers the reference Top-$K$ set if candidate coverage and the stated ranking-margin condition hold.

\section{Experiments}
In this section, we conduct extensive experiments to evaluate the effectiveness of the proposed HC framework. We first detail the experimental setup, followed by a comprehensive comparison with existing baselines, and finally present ablation studies to analyze the contribution of each core component. For a fair comparison, we include the cost required to compute HC's training-dynamics scores in its reported training time cost. An analysis of the proxy's predictive ability and the seed ratio selection are provided in Appendix~\ref{sec:proxy_cv_pearson}. The detailed adaptation of each baseline to the annotation-limited setting is provided in Appendix~\ref{app:baseline_adaptation}. The main paper reports LLaMA-2-7B results, while Qwen2-7B results and implementation details are provided in Appendix~\ref{app:implementation_details} and~\ref{app:qwen_results}. HC uses $\alpha=0.50$, $\beta=0.50$, and $\gamma=0.20$ as the default scoring parameters; all hidden states used in this work are extracted from the last transformer layer at the last non-padding token of the prompt; a detailed sensitivity analysis is provided in Appendix~\ref{app:parameter_sensitivity}.

\subsection{Experimental Setup}

\paragraph{Datasets} We evaluate HC across three instruction-tuning domains: medicine, law, and finance. The training pools are constructed from corpora typically used by domain-specific LLMs~\cite{chen2023meditron70bscalingmedicalpretraining,yang2025fingptopensourcefinanciallarge,yue2023disclawllmfinetuninglargelanguage}. For medicine, we use MedQA, MedQA-4, PubMedQA, MedMCQA, and MMLU-Medical~\cite{jin2020diseasedoespatienthave,jin-etal-2019-pubmedqa,pmlr-v174-pal22a,hendrycks2021measuringmassivemultitasklanguage}. For law, we select task-aligned objective benchmarks from the held-out legal pool: Exam, LED, and LCC. For finance, we evaluate on FPB~\cite{Malo2014GoodDO}, FiQA, and NWGI. All benchmarks use accuracy as the evaluation metric. Detailed dataset statistics, train-test splits, preprocessing rules, and task formats are provided in Appendix~\ref{app:dataset_details}.

\paragraph{Baselines} We compare HC with Full, Random, SPICE~\cite{chang2026spice}, Fisher~\cite{deb2025fishersft}, LESS~\cite{xia2024lessselectinginfluentialdata}, SelectIT~\cite{liu2025selectitselectiveinstructiontuning}, DPP~\cite{zhang2023dppbasedclientselectionfederated}, TSDS~\cite{liu2024tsds}, LEAD~\cite{lin2025leaditerativedataselection}, Superfilter~\cite{li2024superfilteringweaktostrongdatafiltering}, S2L~\cite{yang2024smalltolarge}, Coreset~\cite{sener2018active}, BADGE~\cite{Ash2020Deep} and AIT~\cite{kung2023activeinstructiontuningimproving}. All methods follow the same budget protocol in each domain, and label-dependent baselines are adapted with a two-stage protocol to match our unlabeled setting.

\subsection{Main Results}
\label{sec:main_results}

\subsubsection{Limitation of Existing Methods}

\begin{table}[htbp]
\centering
\small
\renewcommand{\arraystretch}{1.12}
\setlength{\tabcolsep}{1.8pt}
\begin{tabular}{@{}lcccc@{}}
\toprule
\textbf{Benchmark} & \textbf{SPICE} & \textbf{Fisher} & \textbf{LESS} & \textbf{LEAD} \\
\midrule
MedQA       & 33.8 / \underline{35.8} & 32.9 / \underline{35.1} & 28.6 / \underline{36.4} & 18.4 / \underline{36.7} \\
MedQA-4-opt & 39.2 / \underline{41.3} & 38.1 / \underline{40.5} & 29.4 / \underline{40.8} & 29.3 / \underline{42.4} \\
PubMedQA    & 74.8 / \underline{75.0} & \underline{75.8} / 75.2 & 60.2 / \underline{74.4} & \underline{76.4} / 73.8 \\
MedMCQA     & 38.2 / \underline{41.8} & 38.2 / \underline{42.0} & 35.6 / \underline{43.8} & 25.6 / \underline{44.8} \\
MMLU-Med    & \underline{53.9} / 53.7 & \underline{54.2} / 53.2 & 44.6 / \underline{51.6} & 20.5 / \underline{53.8} \\
\midrule
Avg.        & 48.0 / \underline{49.5} & 47.8 / \underline{49.2} & 39.7 / \underline{49.4} & 34.0 / \underline{50.3} \\
\bottomrule
\end{tabular}
\caption{Pseudo-label vs. oracle-label selection in the medical domain. For each method, we select 15\% of the training pool and fine-tune the model under the same setting. Pseudo uses model-generated labels to compute selection signals, while Oracle uses ground-truth labels. Each cell reports Pseudo / Oracle accuracy, and the \underline{underline} indicates the better result within each pair.}
\label{tab:pseudo_oracle_gap}
\vspace{-3mm}
\end{table}

We compare pseudo-label and oracle-label selection for SPICE, Fisher, LESS, and LEAD to measure how much label unavailability degrades label-dependent acquisition. Table~\ref{tab:pseudo_oracle_gap} shows that these methods are highly sensitive to label quality. Replacing pseudo-labels with oracle labels improves the average performance across all four methods: SPICE from 48.0 to 49.5, Fisher from 47.8 to 49.2, LESS from 39.7 to 49.4, and LEAD from 34.0 to 50.3. The gains are especially large for LESS and LEAD, indicating that pseudo-label noise can severely distort loss- and gradient-based acquisition signals. As further illustrated in Figure~\ref{motivation}, label-dependent method (SPICE) and traditional active learning method (BADGE) fail to effectively isolate high-value regions. Instead, their selected samples are scattered across the entire sample space, closely resembling random selection.

\begin{table*}[htbp]
\centering
\small
\renewcommand{\arraystretch}{1.15}
\setlength{\tabcolsep}{1.6pt}
\begin{tabular}{@{}l cccccc c cccc c cccc@{}}
\toprule
\multirow{2}{*}{\textbf{Method}} 
& \multicolumn{6}{c}{\textbf{Medical Domain (30\%-15\%)}} 
& & \multicolumn{4}{c}{\textbf{Legal Domain (20\%-10\%)}} 
& & \multicolumn{4}{c}{\textbf{Financial Domain (30\%-15\%)}} \\
\cmidrule{2-7} \cmidrule{9-12} \cmidrule{14-17}
 & \textbf{MQA} & \textbf{MQA4} & \textbf{PMQA} & \textbf{MMCQA} & \textbf{MMLU} & \textbf{Avg.}
 & & \textbf{Exam} & \textbf{LED} & \textbf{LCC} & \textbf{Avg.}
 & & \textbf{FPB} & \textbf{FIQA} & \textbf{NWGI} & \textbf{Avg.} \\
\midrule
Base      & 22.07 & 29.85 & 55.20 & 30.31 & 37.78 & 35.04 & & 25.60 & 40.80 & 31.20 & 32.53 & & 19.69 & 54.81 & 35.89 & 36.80 \\
Full (100\%) & 39.51 & 44.93 & 78.00 & 46.45 & 56.61 & 53.10 & & 24.20 & 44.40 & 34.00 & 34.20 & & 38.13 & 83.68 & 54.11 & 58.64 \\
\midrule
Random      & 35.21 & 39.71 & 75.08 & 41.87 & 51.30 & 48.63 & & 21.72 & \textbf{42.80} & 41.76 & 35.43 & & 78.46 & 80.67 & 64.56 & \underline{74.56} \\
SPICE       & 34.17 & 40.30 & \underline{75.60} & 40.86 & 55.03 & 49.19 & & 22.40 & 42.40 & 40.80 & 35.20 & & 65.76 & 83.26 & 61.35 & 70.12 \\
Fisher      & 34.49 & 39.75 & 75.40 & 41.19 & \underline{55.24} & 49.21 & & 22.20 & 42.20 & 41.00 & 35.13 & & 67.41 & 81.59 & 32.31 & 60.44 \\
LESS        & 21.13 & 30.40 & 55.20 & 33.28 & 40.32 & 36.07 & & 20.20 & 40.80 & 31.80 & 30.93 & & 28.62 & 33.89 & 30.15 & 30.89 \\
SelectIT    & 17.20 & 23.25 & 62.20 & 20.82 & 37.46 & 32.19 & & 19.00 & 41.40 & \underline{49.60} & 36.67 & & 37.97 & \underline{85.77} & 53.32 & 59.02 \\
DPP         & \underline{36.29} & 40.77 & 73.40 & 41.98 & 53.33 & 49.15 & & 20.60 & 42.00 & 36.00 & 32.87 & & 76.26 & 78.24 & 64.57 & 73.02 \\
TSDS        & 34.41 & 38.57 & 72.40 & 41.17 & 52.28 & 47.77 & & 21.80 & 39.20 & 38.80 & 33.27 & & 79.74 & 76.99 & 65.38 & 74.04 \\
LEAD        & 20.74 & 25.69 & 75.20 & 23.17 & 23.39 & 33.64 & & 20.20 & \underline{42.60} & 42.00 & 34.93 & & \underline{81.64} & 72.38 & 62.37 & 72.13 \\
SuperFilter   & 29.77 & 35.43 & 56.80 & 38.37 & 38.94 & 39.86 & & \textbf{24.80} & 38.00 & 35.80 & 32.87 & & 59.72 & 33.89 & 46.82 & 46.81  \\
S2L      & \underline{36.29} & \underline{41.32} & \textbf{77.00} & \underline{42.29} & 49.63 & \underline{49.31} & & 23.20 & 40.40 & 35.40 & 33.00 & & 76.84 & 74.06 & 65.13 & 72.01 \\
\midrule
Coreset     & 34.72 & 40.53 & 74.20 & 41.55 & 51.22 & 48.44 & & 20.20 & 41.60 & 33.00 & 31.60 & & 59.88 & 70.29 & 55.28 & 61.82 \\
BADGE       & 32.68 & 39.28 & 75.20 & 40.07 & 49.63 & 47.37 & & 21.20 & \textbf{42.80} & 38.40 & 34.13 & & 76.34 & 76.15 & \underline{65.46} & 72.65 \\
AIT     & 22.86 & 31.11 & 55.20 & 33.97 & 44.23 & 37.47 & & 23.80 & 41.60 & 47.80 & \underline{37.73} & & 38.96 & \textbf{86.61} & 55.35 & 60.31 \\
\midrule
\textbf{HC (Ours)} & \textbf{38.19} & \textbf{43.47} & 75.40 & \textbf{43.44} & \textbf{57.19} & \textbf{51.54}$^{\ddagger}$ & & \underline{24.40} & \textbf{42.80} & \textbf{66.76} & \textbf{44.65}$^{\ddagger}$ & & \textbf{84.22} & 83.60 & \textbf{65.77} & \textbf{77.86}$^{\ddagger}$ \\
\bottomrule
\end{tabular}
\caption{Main experimental results (Acc \%) across medical, legal, and financial domains. Full denotes training on the entire available dataset. For budgeted selection methods, the notation $(B\%-K\%)$ denotes first constructing a candidate pool under a $B\%$ annotation budget and then selecting a $K\%$ final training subset. The best method in each column is highlighted in \textbf{bold}, and the second best is \underline{underlined}. For stochastic methods, we report the mean over random seeds 42, 142, 242, 342, and 442. The $t$-tests were conducted on the Average results, where $^{\ddagger}$ indicates $p < 0.01$ against the Random selection baseline.}
\label{tab:main_results}
\vspace{-3mm}
\end{table*}

\begin{figure}[ht]
\centering
\includegraphics[width=\columnwidth]{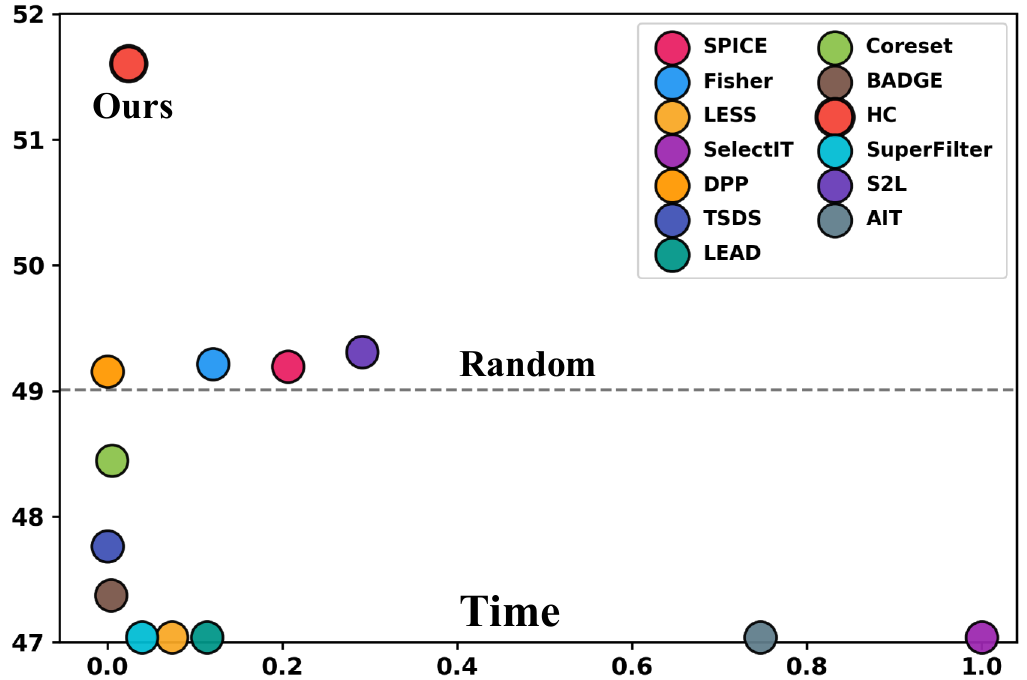}
\caption{Runtime--performance comparison on the medical domain. Each point corresponds to one data-selection method. The $x$-axis reports the time cost over the methods shown in this figure, $x=(t-t_{\min})/(t_{\max}-t_{\min})$. The $y$-axis reports the average accuracy over five medical benchmarks. The dashed horizontal line denotes the accuracy of random. Methods whose average score is below $47$ are clipped onto the $x$-axis rather than plotted at their true $y$-values.}
\label{fig:efficiency_and_score}
\vspace{-3mm}
\end{figure}

\subsubsection{Domin Experiments}

We compare HC against a diverse set of baselines under strict annotation and training constraints. Table~\ref{tab:main_results} reports downstream performance across medical, legal, and financial domains. Overall, HC achieves the strongest average performance among budgeted selection methods in all three domains, while remaining competitive with or outperforming Full under much smaller training budgets. In the medical domain, HC obtains 51.54 $\pm$ 0.16 average accuracy, significantly outperforming Random at 48.63 $\pm$ 0.38 and approaching Full at 53.10. In the legal domain, HC achieves 44.65 $\pm$ 0.60, substantially higher than Random at 35.43 $\pm$ 0.27 and Full at 34.20. In the financial domain, HC reaches 77.86 $\pm$ 0.66, clearly outperforming Random at 74.56 $\pm$ 0.36 and Full at 58.64. These gains over Random are much larger than the corresponding standard deviations, showing that HC consistently improves over random selection across seeds. Across domains, label-dependent methods such as SPICE, Fisher, LESS, SelectIT, Superfilter and LEAD are often unstable under pseudo-label adaptation, while representation-based methods such as AIT, Coreset, and BADGE remain limited by generic diversity objectives. These results show that HC provides a more reliable acquisition signal under annotation and training-budget constraints.

Figure~\ref{fig:efficiency_and_score} compares the runtime-performance trade-off in the medical domain. Representation-based methods are efficient but achieve limited accuracy, while label-dependent methods incur higher costs due to pseudo-labeling or gradient computation. HC achieves the best performance with moderate runtime, demonstrating a balance between selection quality and computational efficiency.

Table~\ref{tab:user_study} shows a post-hoc study comparing HC selected samples against human and LLM judgments of data value. The results demonstrate substantial agreement: the average O@10 across domains reaches 0.70 for a human expert, and 0.53, 0.60, and 0.60 for GPT-5.6, Gemini 3.6 Flash, and Grok 4.5, respectively (prompt details in Appendix~\ref{app:user_study_prompt}).

\begin{table}[htbp]
\centering
\small
\renewcommand{\arraystretch}{1.4}
\setlength{\tabcolsep}{2.0pt}
\begin{tabular}{lcccccc}
\toprule
\multirow{2}{*}{\textbf{Evaluator}}
& \multicolumn{2}{c}{\textbf{Med.}}
& \multicolumn{2}{c}{\textbf{Law}}
& \multicolumn{2}{c}{\textbf{Fin.}} \\
\cmidrule(lr){2-3} \cmidrule(lr){4-5} \cmidrule(l){6-7}
& \textbf{O@10} & \textbf{J@10}
& \textbf{O@10} & \textbf{J@10}
& \textbf{O@10} & \textbf{J@10} \\
\midrule
Human Expert     & 0.70 & 0.54 & 0.60 & 0.43 & 0.80 & 0.67 \\
GPT-5.6          & 0.40 & 0.25 & 0.70 & 0.54 & 0.50 & 0.33 \\
Gemini 3.6       & 0.50 & 0.33 & 0.70 & 0.54 & 0.60 & 0.43 \\
Grok 4.5         & 0.60 & 0.43 & 0.30 & 0.18 & 0.90 & 0.82 \\
\midrule
\textbf{Avg.}    & \textbf{0.55} & \textbf{0.39} & \textbf{0.58} & \textbf{0.42} & \textbf{0.70} & \textbf{0.56} \\
\bottomrule
\end{tabular}
\caption{Human and LLM judge study on high-value sample selection. For each domain, 100 candidate training samples are randomly sampled, and each evaluator selects the 10 samples considered most valuable for instruction tuning. Med., Law, and Fin. denote the medical, legal, and financial domains. O@10 denotes overlap with HC's top-10 selection, computed as $|\mathcal{S}_{\mathrm{HC}}\cap\mathcal{S}_{\mathrm{Eval}}|/10$, and J@10 denotes Jaccard similarity, computed as $|\mathcal{S}_{\mathrm{HC}}\cap\mathcal{S}_{\mathrm{Eval}}|/|\mathcal{S}_{\mathrm{HC}}\cup\mathcal{S}_{\mathrm{Eval}}|$.}
\vspace{-6mm}
\label{tab:user_study}
\end{table}

\subsection{Ablation Study}
\label{sec:ablation}

\begin{table}[htbp]
\centering
\small
\setlength{\tabcolsep}{1.9pt}
\renewcommand{\arraystretch}{1.12}
\begin{tabular}{lcccccc}
\toprule
\textbf{Variant} &
\textbf{MedQA} &
\textbf{MQA4} &
\textbf{PQA} &
\textbf{MMCQA} &
\textbf{MMLU} &
\textbf{Avg.} \\
\midrule
R + E  & 37.94 & 42.97 & 74.80 & 42.86 & 56.51 & 51.02 \\
S + E   & 35.04 & 39.36 & 75.00 & 38.08 & 50.05 & 47.51 \\
C + E & 37.55 & 43.05 & 76.80 & 42.51 & 55.66 & 51.11 \\
D + E    & 38.10 & 42.42 & 76.20 & 42.77 & 55.34 & 50.97 \\
\midrule
P (15\%)  & 37.08 & 41.60 & 74.20 & 42.15 & 54.78 & 49.96 \\
P (30\%)  & 38.26 & 43.99 & 75.20 & 44.32 & 56.19 & 51.59 \\
P + R  & 37.16 & 42.26 & 75.00 & 38.08 & 50.05 & 48.60 \\
P + S   & 37.08 & 41.08 & 75.40 & 42.27 & 53.86 & 49.94 \\
P + C & 37.78 & 41.95 & 75.60 & 42.48 & 53.86 & 50.33 \\
P + D     & 35.74 & 42.97 & 76.00 & 42.24 & 53.23 & 50.04 \\
\specialrule{0.08em}{0.4em}{0.3em}
\textbf{HC}
& \textbf{38.19}
& \textbf{43.47}
& \textbf{75.40}
& \textbf{43.44}
& \textbf{57.19}
& \textbf{51.54} \\
\bottomrule
\end{tabular}
\caption{Component ablation of HC in the medical domain. P denotes PCE, E denotes ECP, R denotes Random, S denotes SPICE, C denotes Coreset, and D denotes DPP. PCE (15\%) directly selects a 15\% subset in Phase 1 using proxy-predicted scores, while PCE (30\%) trains on the full 30\% Phase-1 candidate pool without final pruning. Other variants replace either the candidate construction or the final pruning module.}
\label{tab:hc_component_ablation}
\vspace{-5mm}
\end{table}

\paragraph{Impact of stage 1} We first study whether Progressive Candidate Expansion (PCE) can construct a high-quality candidate pool under the annotation budget. Replacing LLM hidden states with surface-level text features or sentence embeddings weakens candidate recovery under the same budget, showing that target-LLM hidden states better predict dynamics-based sample value. One-shot proxy selection is also weaker than progressive expansion, indicating that iterative refinement improves the proxy as more annotated samples are added. However, using more rounds brings limited gains while increasing retraining cost. Therefore, HC adopts 5 expansion rounds to balance quality and efficiency. Detailed results are reported in Appendix~\ref{app:pce_ablation}. We further replace PCE with existing candidate construction methods while keeping Exact Cartography Pruning fixed. As shown in the upper block of Table~\ref{tab:hc_component_ablation}, Random+ECP, SPICE+ECP, Coreset+ECP, and DPP+ECP all underperform HC. This confirms that ECP cannot fully compensate for a weak candidate pool, and that PCE is necessary for recovering high-value samples under limited annotation.

\paragraph{Impact of stage 2} We next study the role of Exact Cartography Pruning (ECP). Table~\ref{tab:hc_component_ablation} includes two direct PCE variants: PCE (15\%) directly selects a 15\% subset in Phase 1 using proxy-predicted scores, without candidate expansion and pruning, while PCE (30\%) uses the full 30\% Phase-1 candidate pool for training without final pruning. PCE (15\%) performs worse than HC, and PCE (30\%) reaches similar accuracy but uses twice the training data. Other final selectors, including Random, SPICE, Coreset, and DPP, also underperform HC. This shows that ECP improves sample efficiency by pruning the expanded 30\% candidate pool into an effective 15\% subset using exact confidence-variability scores. The results for the financial and legal domains are provided in Appendix~\ref{sec:other_ablation}.

\section{Conclusion}
% This paper presents HC for domain LLM adaptation. Combining Progressive Candidate Expansion with Exact Cartography Pruning, HC outperforms baselines across three domains, cutting annotation and training costs by up to 70--80\% and 85--90\%. Future work will extend HC to multimodal adaptation and reinforcement learning.

This paper presents HC, a novel data selection framework tailored for domain-specific LLM adaptation. Under strict budget constraints, HC leverages target-LLM hidden states to guide progressive candidate expansion, and applies exact cartography pruning to isolate a compact yet highly effective instruction-tuning subset. Extensive experiments across the medical, legal, and financial domains demonstrate that HC consistently outperforms existing budgeted selection methods, achieving superior downstream performance while reducing annotation costs by up to 70--80\% and training costs by up to 85--90\%. Future work will explore extending HC to multimodal adaptation and reinforcement learning scenarios.

\section*{Limitations}
Due to computational constraints, our empirical evaluation is primarily conducted on 7B-scale models, which currently serve as the standard testbed in the instruction-tuning data selection literature. Although the core mechanisms of Hidden Cartographer are theoretically model-agnostic, verifying its selection efficiency on frontier-scale models requires massive distributed training infrastructure. Given that data selection yields more substantial absolute cost reductions as model size increases, a promising direction for future research is to apply HC to trillion-parameter architectures, potentially unlocking even greater computational savings for frontier model alignment.

% \section*{Acknowledgments}

% This document has been adapted
% by Steven Bethard, Ryan Cotterell and Rui Yan
% from the instructions for earlier ACL and NAACL proceedings, including those for
% ACL 2019 by Douwe Kiela and Ivan Vuli\'{c},
% NAACL 2019 by Stephanie Lukin and Alla Roskovskaya,
% ACL 2018 by Shay Cohen, Kevin Gimpel, and Wei Lu,
% NAACL 2018 by Margaret Mitchell and Stephanie Lukin,
% Bib\TeX{} suggestions for (NA)ACL 2017/2018 from Jason Eisner,
% ACL 2017 by Dan Gildea and Min-Yen Kan,
% NAACL 2017 by Margaret Mitchell,
% ACL 2012 by Maggie Li and Michael White,
% ACL 2010 by Jing-Shin Chang and Philipp Koehn,
% ACL 2008 by Johanna D. Moore, Simone Teufel, James Allan, and Sadaoki Furui,
% ACL 2005 by Hwee Tou Ng and Kemal Oflazer,
% ACL 2002 by Eugene Charniak and Dekang Lin,
% and earlier ACL and EACL formats written by several people, including
% John Chen, Henry S. Thompson and Donald Walker.
% Additional elements were taken from the formatting instructions of the \emph{International Joint Conference on Artificial Intelligence} and the \emph{Conference on Computer Vision and Pattern Recognition}.


\bibliography{custom}

\newpage

\appendix

\begin{center}
    {\Large\bfseries Appendices}
\end{center}

\section{Algorithm Implementation}
\label{Algorithm Implementation}
This section presents the pseudocode for the Hidden Cartographer framework. Algorithm~\ref{alg:phase1} describes Progressive Candidate Expansion, which constructs an annotated candidate pool from unlabeled domain data under a limited annotation budget. Algorithm~\ref{alg:phase2} outlines Exact Cartography Pruning, which selects the final instruction-tuning subset from this candidate pool using exact confidence-variability scores.

\begin{algorithm}[h]
\caption{Phase 1: Progressive Candidate Expansion}
\label{alg:phase1}
\textbf{Input}: Unlabeled dataset $\mathcal{U}$, Target LLM $\mathcal{M}_{\theta}$, Proxy network $f_\phi$, Seed size $N_{\mathrm{seed}}$, Round size $N_{\mathrm{round}}$, Active rounds $T$, Parameters $\alpha, \beta, \gamma$ \\
\textbf{Output}: Pool $\mathcal{C}$
\begin{algorithmic}[1]
\STATE $\mathcal{S}_0 \sim \mathrm{Uniform}(\mathcal{U}, N_{\mathrm{seed}})$
\STATE $\mathcal{L}_0 \leftarrow \mathrm{ExpertAnnotate}(\mathcal{S}_0)$
\STATE $\mathcal{C} \leftarrow \mathcal{L}_0$ 
\STATE $\mathcal{U}_{\mathrm{remain}} \leftarrow \mathcal{U} \setminus \mathcal{S}_0$
\FOR{$t = 1$ \TO $T$}
    \FOR{each $x_i \in \mathcal{C}$}
        \STATE $C_i, V_i \leftarrow \mathrm{CalcDynamics}(x_i, y_i, \mathcal{M}_{\theta},\mathcal{C})$
        \STATE $\mathbf{h}_i \leftarrow \mathrm{ExtractHiddenState}(\mathcal{M}_{\theta}, x_i)$
    \ENDFOR
    \STATE $\phi^* \leftarrow \arg\min_{\phi} \sum_{i=1}^{|\mathcal{C}|} \left\| f_{\phi}(\mathbf{h}_i) - (C_i, V_i) \right\|_2^2$
    \FOR{each $x_j \in \mathcal{U}_{\mathrm{remain}}$}
        \STATE $\mathbf{h}_j \leftarrow \mathrm{ExtractHiddenState}(\mathcal{M}_{\theta}, x_j)$
        \STATE $(\hat{C}_j, \hat{V}_j) \leftarrow f_{\phi^*}(\mathbf{h}_j)$
    \ENDFOR
    \STATE $\tilde{C}, \tilde{V} \leftarrow \mathrm{MinMaxScaler}(\hat{C}, \hat{V})$
    \STATE $C_{\mathrm{med}}, V_{\mathrm{med}} \leftarrow \mathrm{Median}(\tilde{C}),\mathrm{Median}(\tilde{V})$
    \FOR{each $x_j \in \mathcal{U}_{\mathrm{remain}}$}
        \STATE $P_{\mathrm{ext}}^{(j)} \leftarrow \left| \tilde{C}_j - C_{\mathrm{med}} \right| + \left| \tilde{V}_j - V_{\mathrm{med}} \right|$
        \STATE $\mathrm{Score}_j \leftarrow \alpha \cdot \tilde{C}_j + \beta \cdot \tilde{V}_j - \gamma \cdot P_{\mathrm{ext}}^{(j)}$
    \ENDFOR
    \STATE $\mathcal{S}_t \leftarrow \mathrm{TopK}(\mathcal{U}_{\mathrm{remain}}, \mathrm{Score}, N_{\mathrm{round}})$
    \STATE $\mathcal{L}_t \leftarrow \mathrm{ExpertAnnotate}(\mathcal{S}_t)$
    \STATE $\mathcal{C} \leftarrow \mathcal{C} \cup \mathcal{L}_t$
    \STATE $\mathcal{U}_{\mathrm{remain}} \leftarrow \mathcal{U}_{\mathrm{remain}} \setminus \mathcal{S}_t$
\ENDFOR
\STATE \textbf{return} $\mathcal{C}$
\end{algorithmic}
\end{algorithm}

\begin{algorithm}[tb]
\caption{Phase 2: Exact Cartography Pruning}
\label{alg:phase2}
\textbf{Input}: Candidate pool $\mathcal{C}$, Target LLM $\mathcal{M}_{\theta}$, Target final size $N_{\mathrm{final}}$, Parameters $\alpha, \beta, \gamma \\$
\textbf{Output}: Dataset $\mathcal{S}_{\mathrm{final}}$
\begin{algorithmic}[1]
\FOR{each $(x_i, y_i) \in \mathcal{C}$}
    \STATE $C_i, V_i \leftarrow \mathrm{CalcDynamics}(x_i, y_i, \mathcal{M}_{\theta}, \mathcal{C})$ 
\ENDFOR
\STATE $\tilde{C},\tilde{V}  \leftarrow \mathrm{MinMaxScaler}(C, V)$
\STATE $C_{\mathrm{med}}, V_{\mathrm{med}} \leftarrow \mathrm{Median}(\tilde{C}), \mathrm{Median}(\tilde{V})$
\FOR{each $x_i \in \mathcal{C}$}
    \STATE $P_{\mathrm{ext}}^{(i)} \leftarrow \left| \tilde{C}_i - C_{\mathrm{med}} \right| + \left| \tilde{V}_i - V_{\mathrm{med}} \right|$
    \STATE $\mathrm{Score}_i \leftarrow \alpha \cdot \tilde{C}_i + \beta \cdot \tilde{V}_i - \gamma \cdot P_{\mathrm{ext}}^{(i)}$
\ENDFOR
\STATE $\mathcal{S}_{\mathrm{final}} \leftarrow \mathrm{TopK}(\mathcal{C}, \mathrm{Score}, N_{\mathrm{final}})$
\STATE \textbf{return} $\mathcal{S}_{\mathrm{final}}$
\end{algorithmic}
\end{algorithm}

\section{Theoretical Assumptions and Proofs}
\label{app:theory}

We provide a deterministic finite-pool analysis of the normalized, median-centered HC score. Our analysis disentangles prediction error, candidate coverage, and changes in the scoring reference. The resulting theoretical guarantees concern cartography-score rankings and sums; they do not identify a universally optimal training subset or imply downstream generation-loss guarantees.

\subsection{Setup and Conventions}

We represent a dataset by its set of example indices. All theoretical results are conditioned on the realized seed, feature extractor, training randomness, and checkpoint trajectories. A trajectory $\tau=(\theta^{(1)},\ldots,\theta^{(E)})$ with $E\ge1$ defines the training raw dynamics $z_i^\tau=(C_i^\tau,V_i^\tau)$ by the probability mean and standard deviation in the main text. Gold responses contain at least one token. Since $0\le p_i^{(e)}\le1$,
\begin{equation}
0\le C_i^\tau\le1,\quad
(V_i^\tau)^2\le C_i^\tau(1-C_i^\tau)\le\tfrac14.
\end{equation}
The variance inequality follows from $(p_i^{(e)})^2\le p_i^{(e)}$. Thus, the raw targets lie in $[0,1]\times[0,\tfrac12]$.

For a nonempty finite scoring pool $\mathcal{P}$ and a coordinate $Z\in\{C,V\}$, write $a_Z=\min_{i\in\mathcal{P}}Z_i$, $b_Z=\max_{i\in\mathcal{P}}Z_i$, and $r_Z=b_Z-a_Z$. Ordinary min--max scaling is $\tilde Z_i=(Z_i-a_Z)/r_Z$ when $r_Z>0$; a constant coordinate is mapped to zero. Scaling is performed independently for the two coordinates. Define
\begin{equation}
(C_{\mathrm{med}}, V_{\mathrm{med}})=\operatorname{median}_{i\in\mathcal{P}}(\tilde C_i, \tilde V_i).
\end{equation}
For an even-sized pool, the median is the arithmetic mean of the two middle order statistics. The score is
\begin{equation}
\begin{aligned}
q_i^{\tau,\mathcal{P}}
={}&\alpha\tilde C_i+\beta\tilde V_i-\gamma\bigl(|\tilde C_i-C_{\mathrm{med}}|
             +|\tilde V_i-V_{\mathrm{med}}|\bigr),
\end{aligned}
\label{eq:app-hc-score}
\end{equation}
with $\alpha,\beta,\gamma\ge0$. We suppress superscripts when the trajectory and pool are clear. Predicted raw coordinates are normalized using their own minimum, maximum, and medians on exactly the same index set as the exact comparator. A fixed total order on example indices resolves all Top-$K$ ties; Top-$K$ always returns exactly $K$ distinct indices.

In an acquisition round, the exact comparator on the unlabeled pool is a counterfactual evaluation of its gold responses under that round's checkpoint trajectory. Those responses are used only to define the analytical comparison; HC does not observe them before querying the annotation oracle. The proxy is fitted to raw $(C,V)$ targets computed on the annotated pool. It does not directly predict a scalar score or pool-normalized targets.

\subsection{From Raw Prediction Error to Score Error}

\noindent\textbf{Lemma 1.}
\textit{For two aligned finite vectors $u,v$ satisfying $\|u-v\|_\infty\le\epsilon$, their minima, maxima, corresponding order statistics, and medians differ by at most $\epsilon$.}

\noindent\textbf{Proof.}
For increasing order statistics, at least $k$ coordinates of $u$ are at most $u_{(k)}$. The corresponding coordinates of $v$ are at most $u_{(k)}+\epsilon$, so $v_{(k)}\le u_{(k)}+\epsilon$. Interchanging $u,v$ gives the reverse inequality. Minima and maxima are the first and last order statistics. The bound also holds for a middle order statistic and for the average of the two middle order statistics, proving median stability. \hfill$\square$

\noindent\textbf{Lemma 2.}
\textit{Let $Z,\hat Z$ be aligned vectors on the same finite pool, with exact range $r_Z>0$ and $\|Z-\hat Z\|_\infty\le\varepsilon_Z$. Using the scaling convention in Section B.1,}
\begin{equation}
\|\tilde Z-\widehat{\tilde Z}\|_\infty
\le
\delta_Z:=\min\{1,2\varepsilon_Z/r_Z\}.
\label{eq:app-minmax-bound}
\end{equation}
\textit{The predicted range need not be positive.}

\noindent\textbf{Proof.}
Write $a=\min Z$, $b=\max Z$, $\hat a=\min\hat Z$, $\hat b=\max\hat Z$, $r=b-a$, and $\hat r=\hat b-\hat a$. Let $u_i=\hat Z_i-Z_i$, $u_a=\hat a-a$, and $u_b=\hat b-b$. Lemma 1 gives $|u_a|,|u_b|\le\varepsilon_Z$. For every index,
\begin{equation}
\begin{aligned}
r(\widehat{\tilde Z}_i-\tilde Z_i)
={}&u_i-(1-\widehat{\tilde Z}_i)u_a
-\widehat{\tilde Z}_i u_b.
\end{aligned}
\end{equation}
For $\hat r>0$, this follows by substituting
$\hat r=r+u_b-u_a$ into the two normalization formulas. For $\hat r=0$, $\widehat{\tilde Z}_i=0$ and $\hat Z_i=\hat a$, and the same identity follows directly. Since $\widehat{\tilde Z}_i\in[0,1]$, the last two terms form a convex combination of endpoint errors. The absolute right-hand side is at most $2\varepsilon_Z$. Both normalized vectors lie in $[0,1]$, giving the additional cap of one. \hfill$\square$

\noindent\textbf{Range dependence.}
The factor two cannot be improved uniformly: take
$Z=(0,r/2,r)$ and $\hat Z=(\epsilon,r/2-\epsilon,r+\epsilon)$ for $0<\epsilon<r/4$. The middle normalized coordinate changes by $2\epsilon/r$. These can be confidence vectors with $r+ \epsilon\le1$.

A nontrivial range condition is necessary for a bound that vanishes with raw error. For $Z=(0.5,0.5,0.5)$ and
$\hat Z=(0.5-\epsilon,0.5,0.5+\epsilon)$, the exact normalized vector is zero, while the predicted vector is $(0,0.5,1)$ for every $\epsilon>0$. For an exact constant coordinate, the universal replacement $\delta_Z=1$ remains valid; no uniformly vanishing bound follows from $\varepsilon_Z$ alone.

\noindent\textbf{Theorem 1.}
\textit{On the same scoring pool, define $\varepsilon_Z,r_Z,\delta_Z$ as above. Then}
\begin{equation}
\begin{aligned}
\|\hat q-q\|_\infty&\le\eta,\\
\eta&=(\alpha+2\gamma)\delta_C
       +(\beta+2\gamma)\delta_V.
\end{aligned}
\end{equation}
\textit{For a constant exact coordinate, use $\delta_Z=1$.}

\noindent\textbf{Proof.}
Lemma 2 bounds the normalized coordinate errors. Lemma 1 then gives
$|\hat C_{\mathrm{med}}-C_{\mathrm{med}}|\le\delta_C$ and
$|\hat V_{\mathrm{med}}-V_{\mathrm{med}}|\le\delta_V$.
For the confidence penalty, the reverse triangle inequality yields
\begin{equation}
\begin{aligned}
&\bigl||\widehat{\tilde C}_i-\hat C_{\mathrm{med}}|
       -|\tilde C_i-C_{\mathrm{med}}|\bigr|\\
&\quad\le
|\widehat{\tilde C}_i-\tilde C_i|
+|\hat C_{\mathrm{med}}-C_{\mathrm{med}}|
\le2\delta_C.
\end{aligned}
\end{equation}
The variability penalty is bounded by $2\delta_V$ in exactly the same way. The linear terms contribute $\alpha\delta_C+\beta\delta_V$, proving the claim. In particular, if both normalized errors are at most $\delta$, then $\|\hat q-q\|_\infty\le(\alpha+\beta+4\gamma)\delta$. This includes movement of both medians. \hfill$\square$

Theorem 1 requires a maximum error on the scored pool. Small training MSE or small expected prediction risk does not alone provide that guarantee. For example, a unit coordinate error on one of $M$ equally weighted examples has average squared error $1/M$, while its maximum error is one.

\subsection{A Sufficient Hidden-State Prediction Condition}

\noindent\textbf{Proposition 1.}
\textit{Fix a trajectory, an annotated set $\mathcal{A}\ne\varnothing$, and a remaining pool $\mathcal{P}\ne\varnothing$. Let $z_i=(C_i,V_i)$ and let $h_i$ be the frozen prompt-only feature. Suppose a map $G$ is $L_G$-Lipschitz from the feature Euclidean norm to the target infinity norm and}
\begin{equation}
\max_{i\in\mathcal{A}\cup\mathcal{P}}
\|z_i-G(h_i)\|_\infty\le\rho.
\end{equation}
\textit{Suppose the fitted proxy $f$ is $L_f$-Lipschitz in the same norms. Define the labeled fitting error and coverage radius}
\begin{equation}
\begin{aligned}
e_{\mathcal{A}}&=\max_{i\in\mathcal{A}}\|f(h_i)-z_i\|_\infty,\\
r_{\mathcal{A},\mathcal{P}}
&=\max_{j\in\mathcal{P}}\min_{i\in\mathcal{A}}\|h_j-h_i\|_2.
\end{aligned}
\end{equation}
\textit{Then}
\begin{equation}
\begin{aligned}
&\max_{j\in\mathcal{P}}\|f(h_j)-z_j\|_\infty\\
&\quad\le e_{\mathcal{A}}
 +(L_f+L_G)r_{\mathcal{A},\mathcal{P}}+2\rho.
\end{aligned}
\label{eq:app-feature-coverage}
\end{equation}

\noindent\textbf{Proof.}
For $j\in\mathcal{P}$, choose a closest $i\in\mathcal{A}$. Insert $f(h_i)$, $z_i$, $G(h_i)$, and $G(h_j)$ between $f(h_j)$ and $z_j$. The five resulting differences are bounded by
$L_f\|h_j-h_i\|_2$, $e_{\mathcal{A}}$, $\rho$,
$L_G\|h_j-h_i\|_2$, and $\rho$, respectively.
Taking the maximum over $j$ proves the result. \hfill$\square$

This is a conditional sufficient bound, not an unconditional theorem that hidden states determine training dynamics. The residual $\rho$ allows different answers or learning trajectories to have similar prompt representations. Smoothness and representation quality must be assessed empirically. Neither the hidden-state dimension nor the annotated sample count alone implies these assumptions or an $N^{-1/2}$ maximum-error rate.

For a fixed feature map and fixed targets, expanding the labeled set cannot increase the nearest-neighbor coverage radius on the remaining pool. However, the fitted proxy, its fitting error and Lipschitz constant, and the target trajectory can all change. Score-guided acquisition need not cover every feature-space region. Consequently, this proposition does not imply monotonic improvement of the proxy across HC rounds and does not require independent sampling of the adaptively acquired labels.

\subsection{Top-K Stability on a Fixed Pool}

\noindent\textbf{Lemma 3.}
\textit{On a finite pool $\mathcal{P}$, let $S^\ast=\operatorname{TopK}_{\mathcal{P}}(q)$ and $\hat S=\operatorname{TopK}_{\mathcal{P}}(\hat q)$, with $1\le K\le|\mathcal{P}|$ and $\|\hat q-q\|_\infty\le\eta$. Then}
\begin{equation}
0\le\sum_{i\in S^\ast}q_i-\sum_{i\in\hat S}q_i
\le2|S^\ast\setminus\hat S|\eta
\le2K\eta.
\end{equation}
\textit{For $K<|\mathcal{P}|$, define
$\Delta_K=q_{(K)}-q_{(K+1)}$. If $\Delta_K>2\eta$, then $\hat S=S^\ast$. For $K=|\mathcal{P}|$, both sets equal the whole pool.}

\noindent\textbf{Proof.}
The lower bound follows because exact Top-$K$ maximizes the score sum over subsets of size $K$. The sets $S^\ast\setminus\hat S$ and $\hat S\setminus S^\ast$ have equal cardinality. For any omitted $i$ and selected $j$, $\hat q_j\ge\hat q_i$, so
$q_i-q_j\le2\eta$. Pair the two difference sets arbitrarily and sum. For recovery, every $i\in S^\ast$ and $j\notin S^\ast$ satisfy
$\hat q_i-\hat q_j\ge\Delta_K-2\eta>0$, so none can swap sides of the boundary. The strict margin rules out boundary ties. \hfill$\square$

\subsection{Progressive Candidate Coverage}

Fix a reference trajectory $\tau_0$ and set
$Q_i=q_i^{\tau_0,\mathcal{U}}$ on the original pool. The reference need not be computed by HC. Let
$G^\ast=\operatorname{TopK}_{\mathcal{U}}(Q)$,
fix a seed $\mathcal{C}_0$, and define
$\mathcal{P}_0=\mathcal{U}\setminus\mathcal{C}_0$.
At round $t$, $\mathcal{P}_t$ denotes the remaining pool just before acquisition,
$q_i^t=q_i^{\tau_t,\mathcal{P}_t}$ is the exact comparator, and $\hat q_i^t$ is the proxy-based score. All scores are aligned by example index. Set
\begin{equation}
\begin{aligned}
\eta_t&\ge\max_{i\in\mathcal{P}_t}|\hat q_i^t-q_i^t|,\\
d_t&=\max_{i\in\mathcal{P}_t}|q_i^t-Q_i|,\\
E_{\mathrm{acq}}&=\max_t(\eta_t+d_t).
\end{aligned}
\end{equation}
Theorem 1 gives a valid $\eta_t$ from that round's raw errors and ranges. By the triangle inequality,
$|\hat q_i^t-Q_i|\le E_{\mathrm{acq}}$ whenever $i$ is eligible at round $t$.

\noindent\textbf{Theorem 2.}
\textit{At each round, acquire the distinct indices
$A_t=\operatorname{Top}_{b_t,\mathcal{P}_t}(\hat q^t)$,
then remove them from the remaining pool. Let
$b=\sum_t b_t$ and
$\mathcal{C}=\mathcal{C}_0\cup\bigcup_t A_t$,
with $|\mathcal{C}_0|+b\le B_{\mathrm{ann}}$.
Write $m=|G^\ast\cap\mathcal{P}_0|$.
If $1\le m\le b<|\mathcal{P}_0|$ and}
\begin{equation}
Q^{\mathcal{P}_0}_{(m)}
-Q^{\mathcal{P}_0}_{(b+1)}>2E_{\mathrm{acq}},
\end{equation}
\textit{then $G^\ast\subseteq\mathcal{C}$. Here,
$Q^{\mathcal{P}_0}_{(j)}$ sorts the fixed reference scores restricted to $\mathcal{P}_0$ in decreasing order.}

\noindent\textbf{Proof.}
Under the common tie-breaking order, the missing reference set
$G^\ast\cap\mathcal{P}_0$ is exactly the first $m$ indices in the restricted reference ranking. Suppose some $i$ in this set is never annotated. It remains eligible in every round. Every acquired $j\in A_t$ satisfies $\hat q_j^t\ge\hat q_i^t$, hence
\begin{equation}
\begin{aligned}
Q_j
&\ge Q_i-2E_{\mathrm{acq}}
\ge Q^{\mathcal{P}_0}_{(m)}-2E_{\mathrm{acq}}
>Q^{\mathcal{P}_0}_{(b+1)}.
\end{aligned}
\end{equation}
Also $Q_i\ge Q^{\mathcal{P}_0}_{(m)}>Q^{\mathcal{P}_0}_{(b+1)}$.
There are at most $b$ indices in $\mathcal{P}_0$ strictly above its $(b+1)$-st score. But all $b$ distinct acquired indices and the omitted index $i$ lie above that threshold, a contradiction. \hfill$\square$

If $m=0$, all reference examples are already in the seed; if $b=|\mathcal{P}_0|$, all examples are annotated. If $m>b$, coverage is impossible for pure Top-$b_t$ acquisition because the acquisition allowance is smaller than the number of missing reference examples. The theorem conditions on the realized seed and actual round-wise errors. It does not claim that uniform seeding, more rounds, or a larger candidate pool alone makes the condition hold.

\noindent\textbf{Exploration queries.}
If arbitrary exploration queries supplement score-guided selection, let $g$ be the total number of distinct score-guided acquisitions. Each score-guided batch must be Top-$g_t$ under a common score vector on its then-eligible pool. Provided the same reference-error bound holds on those pools, the sufficient condition is
\begin{equation}
Q^{\mathcal{P}_0}_{(m)}
-Q^{\mathcal{P}_0}_{(g+1)}>2E_{\mathrm{acq}},
\quad 1\le m\le g<|\mathcal{P}_0|.
\end{equation}
The proof uses the $g$ score-guided queries and a finally omitted reference index in the same counting argument. Exploration queries cannot remove a finally omitted index, so it remains an eligible comparator for every score-guided batch. All queries, including exploration, count against $B_{\mathrm{ann}}$. For a mixed acquisition protocol such as that described in Appendix~\ref{app:implementation_details}, the guarantee uses the actual number of score-guided queries, not the entire acquisition budget.

\subsection{Exact Pruning and Two-Stage Score Regret}

The final annotated pool $\mathcal{C}$ may use a different trajectory and different normalization statistics from the full-pool reference. Define
\begin{equation}
\begin{aligned}
q_i^F&=q_i^{\tau_F,\mathcal{C}},\\
D_F&=\max_{i\in\mathcal{C}}|q_i^F-Q_i|,\\
G_{\mathcal{C}}^\ast&=\operatorname{TopK}_{\mathcal{C}}(Q),\quad
\mathcal{S}_F=\operatorname{TopK}_{\mathcal{C}}(q^F).
\end{aligned}
\end{equation}
Let $m_{\mathrm{miss}}=|G^\ast\setminus\mathcal{C}|$ and
$W=\max_{i\in\mathcal{U}}Q_i-\min_{i\in\mathcal{U}}Q_i$.

\noindent\textbf{Theorem 3.}
\textit{For $1\le K\le|\mathcal{C}|$, let
$s=|G_{\mathcal{C}}^\ast\setminus\mathcal{S}_F|$.
Then}
\begin{equation}
\begin{aligned}
0\le {}&
\sum_{i\in G^\ast}Q_i-\sum_{i\in\mathcal{S}_F}Q_i\\
\le {}&
\underbrace{\sum_{i\in G^\ast}Q_i
-\sum_{i\in G_{\mathcal{C}}^\ast}Q_i}_{A(\mathcal{C})}
+2sD_F\\
\le {}&m_{\mathrm{miss}}W+2KD_F.
\end{aligned}
\label{eq:app-two-stage-regret}
\end{equation}
\textit{Moreover, $s\le\min\{K,|\mathcal{C}|-K\}$.
If $G^\ast\subseteq\mathcal{C}$, $K<|\mathcal{U}|$, and
$Q_{(K)}-Q_{(K+1)}>2D_F$, then $\mathcal{S}_F=G^\ast$.}

\noindent\textbf{Proof.}
Add and subtract $\sum_{i\in G_{\mathcal{C}}^\ast}Q_i$.
Both resulting differences are nonnegative because global and restricted Top-$K$ maximize their respective reference-score sums.
Lemma 3 on $\mathcal{C}$, with the score perturbation $D_F$, bounds the second difference by $2sD_F$.

To bound the omission term, keep all indices in $G^\ast\cap\mathcal{C}$ and add any $m_{\mathrm{miss}}$ other candidate indices to form a size-$K$ set $H\subseteq\mathcal{C}$. Such a set exists since $|\mathcal{C}|\ge K$.
Restricted optimality gives
$\sum_{G_{\mathcal{C}}^\ast}Q\ge\sum_H Q$.
Pair the $m_{\mathrm{miss}}$ missing global indices with the added indices. Each reference-score difference is at most $W$, so
$A(\mathcal{C})\le m_{\mathrm{miss}}W$.
Finally, $s\le K$ and
$|\,\mathcal{S}_F\setminus G_{\mathcal{C}}^\ast\,|=s\le|\mathcal{C}|-K$.

If the reference Top-$K$ is covered, the strict global boundary gap ensures that every covered reference member has a final score strictly above every candidate outside $G^\ast$, by the same $2D_F$ argument as Lemma 3. Hence final pruning recovers $G^\ast$. If $K=|\mathcal{U}|$, feasibility instead forces $\mathcal{C}=\mathcal{S}_F=\mathcal{U}$, so recovery is trivial without a boundary gap. \hfill$\square$

For normalized HC scores, $-2\gamma\le Q_i\le\alpha+\beta$, so $W\le\alpha+\beta+2\gamma$. If Theorem 2's coverage condition holds, the omission term vanishes. Exact pruning is optimal for the final-pool score $q^F$, but its reference-score regret can still be nonzero when $D_F>0$. When $|\mathcal{C}|=K$, there is no within-pool pruning error, as the sharper bound with $s=0$ records.

\subsection{Trajectory and Pool Changes}

\noindent\textbf{Lemma 4.}
\textit{Compare two trajectories with the same number $E$ of paired checkpoints. If}
\begin{equation}
\max_{i,e}|p_i^{(e)}-p_i^{\prime(e)}|\le\zeta,
\end{equation}
\textit{then, on their common example set,
$|C_i-C_i'|\le\zeta$ and $|V_i-V_i'|\le\zeta$.}

\noindent\textbf{Proof.}
The mean bound follows by averaging the absolute probability differences.
For the standard deviation, let
$H=I-\mathbf{1}\mathbf{1}^{\top}/E$, the orthogonal centering projection. For a probability vector $p_i\in\mathbb{R}^E$,
$V_i=\|Hp_i\|_2/\sqrt E$. Since $\|H\|_{2\to2}\le1$, the reverse triangle inequality gives
\begin{equation}
\begin{aligned}
|V_i-V_i'|
&\le\|H(p_i-p_i')\|_2/\sqrt E\\
&\le\|p_i-p_i'\|_2/\sqrt E
\le\zeta.
\end{aligned}
\end{equation}
The argument includes $E=1$, when $H=0$. \hfill$\square$

On the same scoring pool, Theorem 1 converts these raw drift bounds into a score bound by using $\varepsilon_C=\varepsilon_V=\zeta$ and the reference trajectory's coordinate ranges. Pool changes require a separate term. In particular,
\begin{equation}
\begin{aligned}
d_t\le {}&
\max_{i\in\mathcal{P}_t}
|q_i^{\tau_t,\mathcal{P}_t}-q_i^{\tau_0,\mathcal{P}_t}|\\
&+\max_{i\in\mathcal{P}_t}
|q_i^{\tau_0,\mathcal{P}_t}-q_i^{\tau_0,\mathcal{U}}|.
\end{aligned}
\end{equation}
The first term is trajectory drift on a common pool; the second is drift from recalculating minima, maxima, and medians on a restricted pool. The same decomposition applies to $D_F$ with $\mathcal{P}_t$ replaced by $\mathcal{C}$ and $\tau_t$ by $\tau_F$. Neither term is assumed to vanish.

\noindent\textbf{Counterexample: coverage does not ensure ranking invariance.}
Take $\alpha=\beta=0.5$, $\gamma=0.2$, and six raw dynamics pairs:
\begin{center}
\begin{tabular}{c|rrrrrr}
 & $a$ & $b$ & $c$ & $d$ & $e$ & $f$\\
\hline
$C$ & .8 & .6 & .7 & .1 & .9 & .5\\
$V$ & .1 & .2 & .05 & 0 & 0 & .25
\end{tabular}
\end{center}
They are valid two-checkpoint dynamics: the paired probabilities are
$(.7,.9)$, $(.4,.8)$, $(.65,.75)$, $(.1,.1)$,
$(.9,.9)$, and $(.25,.75)$, respectively.
On the full pool, the normalized median-centered scores are
\begin{equation}
\begin{aligned}
(Q_a,Q_b,Q_c)&=(.58,.60,.4425),\\
(Q_d,Q_e,Q_f)&=(-.1975,.3775,.5725).
\end{aligned}
\end{equation}
The unique Top-1 reference is $b$. The candidate pool
$\mathcal{C}=\{a,b,c\}$ contains it. Keeping the raw dynamics unchanged but recalculating normalization and medians on $\mathcal{C}$ yields
\begin{equation}
(q_a^F,q_b^F,q_c^F)
=(17/30,\,4/15,\,11/60).
\end{equation}
Final pruning selects $a$, so coverage alone is insufficient even with no trajectory change. This is why the reference drift $D_F$ cannot be omitted from Theorem 3.

\subsection{What the Median-Centered Penalty Guarantees}

\noindent\textbf{Proposition 2.}
\textit{With normalization parameters and the medians held fixed, write
$g_C(c)=\alpha c-\gamma|c-C_{\mathrm{med}}|$.
Its slopes below and above $C_{\mathrm{med}}$ are
$\alpha+\gamma$ and $\alpha-\gamma$, respectively.
The variability term has slopes $\beta+\gamma$ and $\beta-\gamma$ around $V_{\mathrm{med}}$.}

\noindent\textbf{Proof.}
Below the center, $|c-C_{\mathrm{med}}|=C_{\mathrm{med}}-c$; above it,
$|c-C_{\mathrm{med}}|=c-C_{\mathrm{med}}$. Substitution gives the two affine branches. The variability term is identical with $\alpha$ replaced by $\beta$. \hfill$\square$

Consequently, the default weights give slopes $0.7$ below either median and $0.3$ above it. Within one realized scoring pool, the score remains strictly increasing in each normalized coordinate. The penalty reduces the marginal reward above the medians, but does not guarantee exclusion of extreme or noisy examples. A decreasing upper branch for confidence requires $\gamma>\alpha$, and for variability requires $\gamma>\beta$; equality produces a plateau. These statements hold with the pool statistics fixed and do not assert monotonicity after the whole pool is renormalized.

None of these mathematical facts identifies label noise from $(C,V)$ or proves that a score-maximizing subset minimizes target-domain loss. Such a conclusion would require an additional, independently justified relationship between the surrogate score and the loss after fine-tuning. We therefore use downstream experiments to assess sample utility and do not make that theoretical claim.

\begin{figure*}[htbp]
\centering
\includegraphics[width=\linewidth]{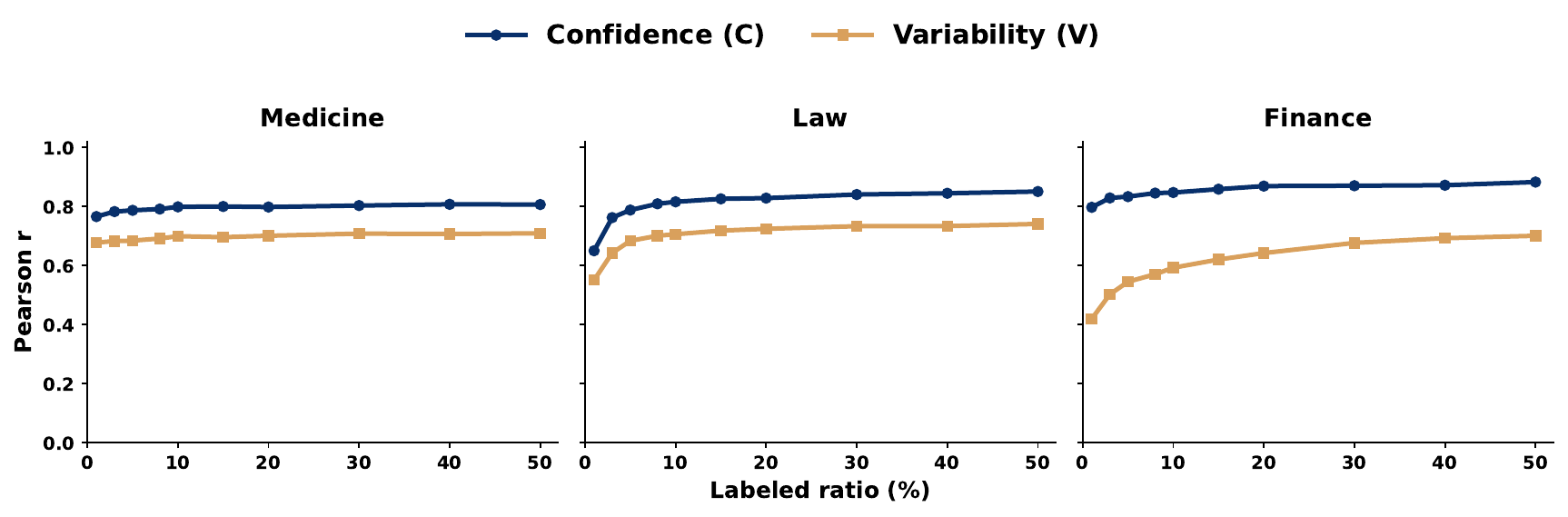}
\caption{Pearson correlation between true and predicted dynamics as a function of the labeled seed size. Each panel is one domain; the two curves are confidence ($C$) and variability ($V$).}
\label{fig:proxy_cv_pearson}
\end{figure*}

\section{Proxy Prediction of Confidence and Variability}
\label{sec:proxy_cv_pearson}

A central assumption of HC is that a small labeled seed is sufficient to train a proxy capable of accurately estimating the true training dynamics. To verify this, we train the HC proxy using prompt-only hidden states to predict the exact confidence ($C$) and variability ($V$). We then report the Pearson correlation coefficient ($r$) on held-out examples as the labeled data ratio increases from 1\% to 50\%.

Figure~\ref{fig:proxy_cv_pearson} demonstrates that the proxy effectively captures both training dynamics. Across the medical, legal, and financial domains, both correlations increase rapidly with a small number of labeled examples before saturating. With a 5\% seed, the Pearson $r$ for $C$ and $V$ reaches approximately $0.79$ and $0.68$ in medicine, $0.79$ and $0.68$ in law, and $0.83$ and $0.55$ in finance, respectively. Increasing the seed from 1\% to 5\% yields substantial gains—particularly for the legal domain and for financial variability—whereas further expanding the seed to 30\% or 50\% produces only marginal improvements. This performance elbow at the 5\% threshold justifies our seed size selection for Phase~1 of HC. Specifically, a 1\%--3\% seed underfits $V$ in the legal and financial domains, causing the subsequent active expansion to rely on an unreliable initial ranking. Conversely, increasing the seed to 10\% or 15\% yields negligible improvements in correlation while unnecessarily expending a significant portion of the annotation budget on random sampling rather than proxy-guided acquisition. Therefore, the 5\% seed represents an optimal operating point: it ensures the proxy is sufficiently accurate to predict $C$ and $V$, while preserving the majority of the Phase-1 budget for targeted expansion.

\section{Experimental Details}
\label{app:experimental_details}

\subsection{Dataset}
\label{app:dataset_details}

This subsection details how we construct the training and evaluation data for the medical, financial, and legal domains. All datasets are converted into a unified instruction-tuning format with three core fields: \texttt{instruction}, \texttt{input}, and \texttt{output}. The \texttt{instruction} field describes the task, the \texttt{input} field contains the question, context, or instance, and the \texttt{output} field stores the answer. For objective tasks, we additionally keep a \texttt{choices} field when candidate options are available, to enable evaluation via option scoring rather than unconstrained generation.

\paragraph{Medical domain.}
The medical training pool is constructed from MedQA~\citep{jin2020diseasedoespatienthave}, PubMedQA~\citep{jin-etal-2019-pubmedqa}, and MedMCQA~\citep{pmlr-v174-pal22a}. For each source, we use its training split and randomly sample 20\% of the records with seed 42. The sampled records are then converted into the unified SFT format and concatenated into a single training pool. MedQA and MedMCQA are formatted as multiple-choice medical exam questions, while PubMedQA is formatted as a biomedical yes/no/maybe question-answering task. The final medical training pool contains 80,852 samples, including 2,035 from MedQA, 42,253 from PubMedQA, and 36,564 from MedMCQA. For evaluation, we use five medical benchmarks: MedQA with five options, MedQA with four options, PubMedQA, MedMCQA, and MMLU-Medical~\citep{hendrycks2021measuringmassivemultitasklanguage}. Their test sizes are 1,273, 1,273, 500, 4,183, and 945, respectively. Each benchmark uses a task-specific instruction consistent with its answer format.

\paragraph{Financial domain.}
The financial training pool is built from six FinGPT instruction-tuning datasets~\citep{yang2025fingptopensourcefinanciallarge}: sentiment classification, headline classification, financial relation extraction (FinRED), named entity recognition, FiQA question answering, and FinEval. We load the training split of each source, normalize every instance into the \texttt{instruction}/\texttt{input}/\texttt{output} schema, and sample 20\% of each dataset with seed 42. This produces 41,032 training samples, including 15,354 sentiment samples, 16,432 headline samples, 5,511 FinRED samples, 102 NER samples, 3,422 FiQA-QA samples, and 211 FinEval samples. For evaluation, we use FPB~\citep{Malo2014GoodDO}, FiQA, and NWGI. The test sets contain 1,209 FPB samples, 239 FiQA samples, and 4,047 NWGI samples. All financial sentiment labels are mapped into the shared label space {\texttt{negative}, \texttt{neutral}, \texttt{positive}}. To avoid train-test leakage, we normalize the input text via lowercasing and whitespace normalization, and remove test examples whose normalized inputs overlap with the training pool.

\paragraph{Legal domain.}

The legal training pool is constructed from DISC-Law-SFT~\citep{yue2023disclawllmfinetuninglargelanguage}. Since the original dataset contains both objective and open-ended legal tasks, we filter it to keep only tasks aligned with our evaluation setting: legal examination questions (\texttt{exam}), legal event detection (\texttt{leg\_eve\_detec}), and legal case classification (\texttt{leg\_case\_cls}). Open-ended legal question answering and legal element extraction tasks are removed. For examination questions, we parse the answer into option labels such as A/B/C/D or comma-separated multi-answer labels, and remove explanation text from the target output. For event detection and case classification, we keep short label outputs and discard overly long generation-style targets. The resulting legal training pool contains 11,204 samples, including 2,908 examination samples, 4,198 legal event detection samples, and 4,098 legal case classification samples. The legal evaluation set is built as a task-aligned objective benchmark with 1,500 samples: 500 examination, 500 legal event detection, and 500 legal case classification examples. Training-overlapping examples are excluded when constructing the held-out legal benchmark. 

\subsection{Baseline Adaptation}
\label{app:baseline_adaptation}

This subsection describes how we adapt existing data selection and active learning baselines to our annotation-limited setting. In their original formulations, many baselines assume that labels are already available for the entire training pool. This assumption is incompatible with our target scenario, where the initial domain pool is unlabeled and only a limited fraction of samples can be annotated by experts. Therefore, all baselines are evaluated under the same budget protocol as HC.

\paragraph{Budget protocol.}
For each domain, we employ a two-stage budget protocol. The first stage constructs an annotated candidate pool under the annotation budget, and the second stage selects the final training subset from this pool. In the medical and financial domains, the protocol constructs a 30\% candidate pool followed by a 15\% final training subset. In the legal domain, the protocol constructs a 20\% candidate pool followed by a 10\% final training subset. Thus, the second stage always selects exactly half of the annotated candidate pool. The same random seed is used across methods unless otherwise specified.

\paragraph{Full and Random.}
The Full baseline fine-tunes the model on the entire available training pool. It does not adhere to the training budget constraint and is included as an empirical upper bound. The Random baseline shuffles the training pool using the specified seed and directly selects a subset matching the final budget (i.e., 15\% for medicine and finance, and 10\% for law). Random selection does not rely on labels, model scores, hidden states, or training dynamics.

\paragraph{Label-dependent baselines.}
SPICE, Fisher, LESS, SelectIT, LEAD, Superfilter, S2L, and AIT~\cite{xia2024lessselectinginfluentialdata,deb2025fishersft,lin2025leaditerativedataselection,chang2026spice,liu2025selectitselectiveinstructiontuning,li2024superfilteringweaktostrongdatafiltering,yang2024smalltolarge,kung2023activeinstructiontuningimproving} require output-conditioned signals such as losses, gradients, influence scores, quality scores, response-loss trajectories, or response likelihoods under prompt perturbations. Since true labels are unavailable before annotation, we adapt these methods with a two-stage candidate-pool protocol. In Stage~1, each method operates on the unlabeled pool using pseudo-labels or pseudo-responses generated by the base LLM. The method then selects the first-stage candidate pool, which is treated as the subset sent to experts. In Stage~2, true outputs are revealed only inside this candidate pool, simulating expert annotation, and the same method selects the final subset from the annotated pool.

For SPICE, the implementation first obtains pseudo-labels for each sample. If the input contains explicit candidate options, the pseudo-label is chosen by candidate log-likelihood scoring; otherwise, it is generated by the model. The method then constructs compact gradient proxies from the pseudo-label loss and selects examples using the SPICE Fisher/conflict criterion. In Stage~2, the gradients are recomputed using the true output of samples inside the expert pool.
Fisher follows the same pseudo-label to true-label adaptation. Stage~1 builds pseudo-label gradient proxies and greedily selects samples by maximizing the log-determinant Fisher information gain. Stage~2 recomputes gradient proxies with true outputs inside the expert pool and applies the same Fisher selection rule.

LESS is adapted by computing pseudo-label gradient features for the unlabeled pool and ranking samples by low-rank gradient similarity to a target gradient direction. When no external target set is specified, the implementation uses the mean training gradient as the target direction. In Stage~2, gradients inside the expert pool are recomputed with true outputs before final selection.

SelectIT is adapted as a two-stage scoring baseline. In Stage~1, the model first generates a pseudo-response for each unlabeled sample. It then uses a reflection-style prompt to score the sample from 1 to $K$ and selects the highest-scoring examples as the candidate pool. In Stage~2, the pseudo-response is replaced by the true output, and SelectIT rescoring is applied within the expert pool.

LEAD is implemented as a one-shot adaptation of its iterative data utility idea. Stage~1 uses pseudo-label losses to compute a static IDU-style utility score and constructs the candidate pool. The default implementation uses cluster-aware selection, where gradient proxies are clustered and selection quotas are allocated according to cluster utility. Stage~2 replaces pseudo-label losses with true-output losses inside the candidate pool and selects the final subset.

Superfilter originally scores already-labeled instruction--response pairs with a frozen weak language model, using instruction-following difficulty $\mathrm{IFD}=\mathrm{PPL}(y\mid x)/\mathrm{PPL}(y)$ and retaining examples with $\mathrm{IFD}<1$. In our setting, Stage~1 first generates a pseudo-response from the instruction and input with the target LLM, then scores that pair with a frozen GPT-2 (124M) scorer; the candidate pool is the highest-IFD subset among finite $\mathrm{IFD}<1$ examples. Stage~2 reveals true outputs only inside this pool and recomputes IFD with the same frozen scorer. To match a hard training budget, if fewer than the required number of examples satisfy $\mathrm{IFD}<1$, we backfill from finite $\mathrm{IFD}\ge 1$ examples nearest the threshold; this backfill is an explicit budget adaptation of the original strict filter rather than a standard Superfilter practice. The weak scorer is never trained.

S2L originally clusters examples by the loss trajectory of a small reference model trained on labeled responses. We keep that clustering rule, but Stage~1 trains the reference only on target-LLM pseudo-responses for the unlabeled pool. Each example is represented by its response-token mean cross-entropy at successive reference-model states (initialization, every 500 optimizer updates, and the final state). Selection uses Euclidean $k$-means on these raw trajectories, with source-balanced uniform sampling inside clusters and a tiny-cluster fallback, matching the official implementation. Stage~2 reloads a fresh reference from the same initialization and retrains it only on the expert pool, using true outputs as labels, then repeats trajectory clustering to obtain the final subset. The default reference is Pythia-70M with full-parameter training; existing cartography fields $(C,V)$ and gold outputs are never used as Stage~1 features.

AIT originally selects tasks by prompt uncertainty: the mean absolute change in the likelihood of a fixed response when instruction words are randomly dropped. We treat each SFT row as one selection unit and do not iterate model retraining. Stage~1 generates a pseudo-response with the frozen target LLM, then scores prompt uncertainty on that fixed response under 20 word-drop perturbations of the instruction (drop probability $0.2$), keeping the input and answer template unchanged. Probability is the length-normalized geometric token probability $\exp(-\overline{\mathrm{NLL}})$ on answer tokens, as in the official mean cross-entropy implementation. Stage~2 substitutes true outputs only inside the candidate pool and recomputes the same perturbation scores. Ranking is by prompt uncertainty in descending order. No scoring model is trained at selection time.

\paragraph{Representation and diversity baselines.}
Coreset, BADGE, DPP, and TSDS~\cite{sener2018active,Ash2020Deep,zhang2023dppbasedclientselectionfederated,liu2024tsds} are adapted as unlabeled or weakly label-dependent baselines. For these methods, the first-stage features are extracted only from the prompt side, i.e., the instruction and input, without using the gold output. The implementation caches target LLM hidden-state embeddings from the last prompt token and projects them to a lower-dimensional feature space. It also records entropy at the last prompt position as an uncertainty signal when needed.

Coreset uses $k$-center greedy selection over prompt hidden-state embeddings. In the two-stage setting, it first selects the candidate pool by maximizing feature coverage. In Stage 2, when true outputs are available inside the candidate pool, the method builds true-output gradient proxies and applies $k$-center selection again within the annotated pool.

BADGE uses an entropy-weighted $k$-means++ procedure over hidden-state embeddings. The first center is selected by maximum entropy by default, and later centers are sampled according to a product of distance and uncertainty. This adapts BADGE to LLM instruction data without requiring full labels before annotation. In Stage 2, the method is rerun inside the candidate pool, using true-output losses as the uncertainty signal since expert labels are now available.

DPP is implemented as a diversity-based unlabeled selection method. Each sample is represented by a normalized prompt hidden-state embedding, and selection is performed using a scalable pool-wise greedy log-determinant approximation. By default, DPP uses pure diversity without label-dependent quality weights. To match the two-stage budget protocol, we first select the candidate pool and then rerun DPP inside this pool to obtain the final subset.

TSDS uses prompt hidden-state embeddings to estimate task-specific sampling probabilities. When no external query set is provided, the method samples a small self-query set from the candidate pool. It then assigns probability mass to nearest neighbors of the query examples and selects samples with the highest probability. As with DPP, TSDS is run once to construct the candidate pool and again within that pool to match the final training budget.

\paragraph{Fairness of adaptation.}

All adapted baselines use the same input pool, random seed, base model, and budget protocol within each domain. For label-dependent methods, true outputs are never used before the first-stage candidate pool is selected. They are only revealed inside the candidate pool to simulate expert annotation. For representation-based methods, prompt-side hidden states are extracted without using gold outputs. This ensures that all baselines operate under the same annotation constraint as HC.

\subsection{Implementation Details}
\label{app:implementation_details}

All main experiments are conducted on a single NVIDIA RTX PRO 6000 GPU with 96GB of memory. Unless otherwise specified, all experiments use the same preprocessing pipeline, budget protocol, and evaluation scripts across methods within each domain.

\paragraph{LoRA fine-tuning.}
For all downstream fine-tuning experiments, we use AdamW with $\beta_1=0.9$, $\beta_2=0.95$, $\epsilon=1\times10^{-5}$, a learning rate of $2\times10^{-5}$, a weight decay of 0.1, and an effective batch size of 64. Models are trained for 3 epochs with a cosine learning rate schedule and a 10\% warmup. The final learning rate decays to 10\% of the peak learning rate. LoRA uses rank $r=16$, a scaling factor of 32, a dropout of 0.05, and is applied to the attention and MLP projection modules. We use mixed precision and enable gradient checkpointing when needed for memory efficiency.

\paragraph{Training dynamics computation.}
To compute training dynamics, we train the target model for 5 epochs and save one LoRA checkpoint after each epoch. For each annotated sample, we evaluate the probability assigned to its gold output under every checkpoint. Confidence is computed as the mean gold-output probability across checkpoints, and variability is computed as the standard deviation. Both probability and loss are computed only over output tokens, while instruction and input tokens are masked out. Therefore, the dynamics reflect the model's learning behavior on the target answer rather than prompt reconstruction.

\paragraph{HC proxy and candidate expansion.}
HC starts from a 5\% randomly annotated seed set. The sample value proxy is trained on this seed set using hidden-state features extracted from the last token of the last transformer layer in the target LLM. The candidate pool is then expanded through active rounds, where each round adds another 5\% of the training pool. 

For the medical and financial domains, HC constructs a 30\% annotated candidate pool and selects the final 15\% training subset. For the legal domain, HC uses a 20\% candidate pool and selects the final 10\% subset, following the smaller legal-domain budget used in the main experiments. After candidate expansion, HC recomputes exact confidence and variability within the annotated candidate pool and applies exact cartography pruning. The final score uses $\alpha=0.50$, $\beta=0.50$, and $\gamma=0.20$ to balance confidence, variability, and the extreme-sample penalty.

\subsection{Other Results}
\label{app:qwen_results}

To verify that our conclusions generalize beyond LLaMA-2-7B, we additionally evaluate all methods using Qwen2-7B as the base model. The experimental protocol, data splits, budget settings, and evaluation scripts remain identical to those in the main experiments. Table~\ref{tab:qwen_results} reports the results across the medical, legal, and financial domains.

\begin{table*}[htbp]
\centering
\small
\renewcommand{\arraystretch}{1.15}
\setlength{\tabcolsep}{1.6pt}
\begin{tabular}{@{}l cccccc c cccc c cccc@{}}
\toprule
\multirow{2}{*}{\textbf{Method}}
& \multicolumn{6}{c}{\textbf{Medical Domain (30\%-15\%)}}
& & \multicolumn{4}{c}{\textbf{Legal Domain (20\%-10\%)}}
& & \multicolumn{4}{c}{\textbf{Financial Domain (30\%-15\%)}} \\
\cmidrule{2-7} \cmidrule{9-12} \cmidrule{14-17}
& \textbf{MQA} & \textbf{MQA4} & \textbf{PMQA} & \textbf{MMCQA} & \textbf{MMLU} & \textbf{Avg.}
& & \textbf{Exam} & \textbf{LED} & \textbf{LCC} & \textbf{Avg.}
& & \textbf{FPB} & \textbf{FIQA} & \textbf{NWGI} & \textbf{Avg.} \\
\midrule
Base      & 49.41 & 55.46 & 37.00 & 53.67 & 71.53 & 53.41 & & 49.40 & 41.80 & 28.60 & 39.93 & & 76.92 & 64.85 & 64.93 & 68.90 \\
Full (100\%) & 54.67 & 59.86 & 78.80 & 58.62 & 72.70 & 64.93
& & 47.60 & 43.20 & 71.60 & 54.13
& & 86.44 & 87.45 & 65.97 & 79.95 \\
\midrule
Random   & \textbf{54.12} & 56.23 & 76.20 & 55.52 & \textbf{73.12} & \underline{63.04}
& & 48.00 & 42.80 & 69.60 & 53.47
& & \underline{81.47} & 82.01 & 65.01 & 76.16 \\
SPICE    & 52.95 & 50.21 & 75.20 & 49.92 & 71.85 & 60.03
& & 48.60 & \textbf{43.00} & 48.40 & 46.67
& & 64.68 & 76.57 & 62.12 & 67.79 \\
Fisher   & 53.26 & 53.52 & 75.40 & 48.52 & 71.75 & 60.49
& & 48.00 & \textbf{43.00} & 47.40 & 46.13
& & 62.78 & 77.41 & 61.87 & 67.35 \\
LESS     & 49.02 & 56.72 & 55.20 & 54.39 & 71.75 & 57.42
& & \underline{50.60} & 41.20 & 70.20 & 54.00
& & 76.34 & 44.35 & 65.33 & 62.01 \\
SelectIT & 49.65 & 56.40 & 65.20 & 55.94 & 72.06 & 59.85
& & 46.20 & 39.80 & 29.60 & 38.53
& & 78.66 & 48.79 & 65.46 & 64.30 \\
DPP      & 51.20 & 49.39 & 74.40 & \underline{57.57} & 72.28 & 60.97
& & 48.60 & 40.00 & 53.80 & 47.47
& & 79.90 & 76.57 & \underline{65.70} & 74.06 \\
TSDS     & 50.52 & 54.83 & \textbf{77.20} & \textbf{57.73} & 68.69 & 61.79
& & 48.00 & 41.80 & 57.80 & 49.20
& & 79.82 & 84.10 & 65.01 & 76.31 \\
LEAD     & 25.06 & 31.19 & \textbf{77.20} & 34.52 & 26.03 & 38.80
& & 49.60 & 41.20 & \underline{82.80} & \underline{57.87}
& & 80.48 & 85.36 & 64.15 & \underline{76.66} \\
SuperFilter   & 53.02 & \textbf{58.44} & 33.80 & 56.18 & 72.06 & 54.70 & & 49.60 & 41.20 & 40.40 & 43.73 & & 68.57 & 85.77 & 61.11 & 71.82  \\
S2L      & 52.67 & 55.78 & \underline{77.00} & 57.14 & 71.64 & 62.85 & & 48.40 & 42.20 & 57.00 & 49.20 & & 77.25 & 77.41 & 65.18 & 73.28 \\
\midrule
Coreset  & \underline{53.97} & 48.47 & 72.60 & 53.73 & 71.80 & 60.11
& & 47.60 & 40.60 & 40.20 & 42.80
& & 54.92 & \textbf{87.01} & 58.96 & 66.96 \\
BADGE    & 53.81 & 55.54 & \underline{77.00} & \underline{57.57} & 71.64 & 63.11
& & 47.80 & 41.40 & 62.40 & 50.53
& & 69.98 & 81.17 & 61.80 & 70.98 \\
AIT      & 51.77 & 56.40 & 72.80 & 56.01 & 71.96 & 61.79 & & 50.40 & 38.00 & 80.80 & 56.40 & & 77.34 & 64.44 & \textbf{66.10} & 69.29 \\
\midrule
\textbf{HC (Ours)}
& \textbf{54.12} & \underline{57.50} & \textbf{77.20} & 56.39 & \underline{72.70} & \textbf{63.58}$^{\ddagger}$
& & \textbf{51.40} & \underline{42.80} & \textbf{83.60} & \textbf{59.27}$^{\ddagger}$
& & \textbf{84.53} & \underline{86.61} & 65.26 & \textbf{78.80}$^{\ddagger}$ \\
\bottomrule
\end{tabular}
\caption{Qwen2-7B results (Accuracy \%) across medical, legal, and financial domains. Full denotes training on the entire available dataset. For budgeted methods, $(B\%-K\%)$ denotes constructing a $B\%$ candidate pool and selecting a $K\%$ final training subset. The best method in each column is highlighted in \textbf{bold}, and the second best is \underline{underlined}. The $t$-tests were conducted on the Average results, where $^{\ddagger}$ indicates $p < 0.01$ against Random.}
\label{tab:qwen_results}
\end{table*}

\subsection{User Study Prompt}
\label{app:user_study_prompt}

This subsection provides the prompt used in the study involving human and LLM judges. For each domain, we randomly sample 100 candidate training examples and ask the evaluator to select the 10 most valuable samples for instruction tuning.

\begin{quote}
\small
You are evaluating data for private-domain LLM instruction tuning.

You will receive 100 candidate training samples from the \{domain\} domain. Each sample contains an instruction, input, and output. Your task is to select the 10 samples that would be most valuable for fine-tuning a domain-specific LLM.

Selection criteria:

1. Prefer samples that are domain-relevant and informative.

2. Prefer samples that teach useful domain knowledge, reasoning, or decision boundaries.

3. Avoid samples that are too trivial, noisy, redundant, or poorly formed.

4. Judge only sample quality for instruction tuning. Do not try to match any hidden algorithm.

Return your answer as JSON only:

\{
\quad ``selected\_ids'': [id1, id2, ..., id10],

\quad ``brief\_reason'': ``one short sentence explaining the selection criteria''
\}

Candidate samples:
\end{quote}

\section{Additional Results of Ablation Study}
\label{app:other_ablation}

\subsection{Additional Results of PCE}
\label{app:pce_ablation}
This section presents detailed results from the ablation study on Progressive Candidate Expansion (PCE). We evaluate whether the first stage of HC can effectively recover high-value samples under a fixed annotation budget. All experiments are conducted in the medical domain using the same 30\% candidate budget. The true high-value set is defined as the top 15\% of samples, as ranked by the exact HC score derived from confidence and variability. We report three metrics: \textbf{Coverage} measures the fraction of the true high-value set successfully recovered by the candidate pool; \textbf{Density} measures the proportion of high-value samples within the constructed candidate pool; and \textbf{Mean} reports the average exact HC score of the pooled samples. Higher values across these metrics indicate that the candidate pool better captures the high-value region targeted by HC during final pruning.

\begin{table}[t]
\centering
\small
\setlength{\tabcolsep}{6pt}
\renewcommand{\arraystretch}{1.12}
\begin{tabular*}{\columnwidth}{@{\extracolsep{\fill}}lccc@{}}
\toprule
\textbf{Method} & \textbf{Cov. $\uparrow$} & \textbf{Dens. $\uparrow$} & \textbf{Mean $\uparrow$} \\
\midrule
\multicolumn{4}{@{}l}{\textbf{(a) Replace hidden-state features}} \\
\cmidrule(lr){1-4}
Text + Metadata    & 56.03 & 28.02 & 0.4223 \\
Sentence Embedding & 55.26 & 27.63 & 0.4213 \\
\midrule
\multicolumn{4}{@{}l}{\textbf{(b) Vary active expansion rounds}} \\
\cmidrule(lr){1-4}
1 Round (Direct Proxy) & 61.58 & 30.79 & 0.4280 \\
10 Rounds              & 68.26 & 34.13 & 0.4321 \\
100 Rounds             & \textbf{69.23} & \textbf{34.61} & \textbf{0.4331} \\
\midrule
\textbf{HC (5 Rounds)} & 67.68 & 33.84 & 0.4313 \\
\bottomrule
\end{tabular*}
\caption{Ablation of feature sources and expansion rounds on the medical domain. Block (a) replaces HC's hidden-state features, while Block (b) varies the number of active rounds under the same 30\% candidate budget.}
\label{tab:active_candidate_expansion}
\end{table}

Table~\ref{tab:active_candidate_expansion} demonstrates the critical role of hidden-state features in candidate construction. Replacing hidden states with surface-level Text + Metadata or Sentence Embedding features substantially degrades both coverage and density, indicating that generic features fail to capture the training-dynamics signals required to identify high-value instruction-tuning samples. The results also highlight the impact of active expansion rounds. A one-shot Direct Proxy selection from the initial seed set performs noticeably worse than progressive expansion. While increasing the number of rounds steadily improves candidate quality, the marginal gains diminish rapidly. Compared to the 5-round default used in HC, expanding to 10 or 100 rounds yields minimal improvement despite significantly increasing proxy retraining costs. Consequently, HC adopts 5 expansion rounds to strike a practical balance between candidate quality and computational efficiency.

\begin{table}[htbp]
\centering
\small
\setlength{\tabcolsep}{3.0pt}
\renewcommand{\arraystretch}{1.12}
\begin{tabular}{lcccc}
\toprule
\textbf{Variant} &
\textbf{FPB} &
\textbf{FiQA} &
\textbf{NWGI} &
\textbf{Avg.} \\
\midrule
Random + ECP & 81.37 & 79.74 & 65.28 & 75.46 \\
SPICE + ECP & 82.80 & 80.75 & 60.10 & 74.55 \\
Coreset + ECP & 83.46 & 80.75 & 66.05 & 76.75 \\
DPP + ECP & 83.54 & 78.52 & 64.32 & 75.46 \\
\midrule
PCE (15\%) & 83.37 & 83.26 & 65.75 & 77.46 \\
PCE (30\%) & 84.70 & 86.61 & 65.51 & 78.94 \\
PCE + Random & 82.55 & 82.52 & 65.01 & 76.69 \\
PCE + SPICE & 81.31 & 82.01 & 64.49 & 75.94 \\
PCE + Coreset & 82.63 & 83.26 & 64.62 & 76.84 \\
PCE + DPP & 81.31 & 82.85 & 65.04 & 76.40 \\
\midrule
\textbf{HC}
& \textbf{84.22}
& \textbf{83.60}
& \textbf{65.77}
& \textbf{77.86} \\
\bottomrule
\end{tabular}
\caption{Component ablation of HC in the financial domain. PCE (15\%) directly selects a 15\% subset in Phase 1 using proxy-predicted scores, while PCE (30\%) trains on the full 30\% Phase-1 candidate pool without final pruning. Other variants replace either the candidate construction or the final pruning module.}
\label{tab:hc_component_ablation_fin}
\vspace{-5mm}
\end{table}

\subsection{Additional domin Results}
\label{sec:other_ablation}

As demonstrated in Tables~\ref{tab:hc_component_ablation_fin} and~\ref{tab:hc_component_ablation_law}, substituting PCE with Random, SPICE, Coreset, or DPP generally degrades average performance. Similarly, replacing ECP with these generic baselines compromises the quality of the final subset. Directly retaining the full Phase-1 pool (PCE 30\%/20\%) or pruning it solely based on proxy scores (PCE 15\%/10\%) also yields inferior results compared to the complete HC pipeline. These trends align with the findings from the medical domain ablation: both PCE and ECP are indispensable, and neither candidate construction nor final pruning can be replaced by generic diversity-based or random modules without incurring a performance loss.

\begin{table}[htbp]
\centering
\small
\setlength{\tabcolsep}{3.0pt}
\renewcommand{\arraystretch}{1.12}
\begin{tabular}{lcccc}
\toprule
\textbf{Variant} &
\textbf{Exam} &
\textbf{LED} &
\textbf{LCC} &
\textbf{Avg.} \\
\midrule
Random + ECP & 21.40 & 43.20 & 65.40 & 43.33 \\
SPICE + ECP & 23.00 & 42.80 & 36.00 & 33.93 \\
Coreset + ECP & 23.40 & 40.40 & 55.80 & 39.87 \\
DPP + ECP & 21.20 & 43.40 & 56.60 & 40.40 \\
\midrule
PCE (10\%) & 25.40 & 42.80 & 62.20 & 43.47 \\
PCE (20\%) & 23.40 & 43.20 & 57.20 & 41.27 \\
PCE + Random & 25.60 & 39.60 & 56.40 & 40.53 \\
PCE + SPICE & 23.60 & 42.20 & 46.00 & 37.27 \\
PCE + Coreset & 25.40 & 42.40 & 50.60 & 39.47 \\
PCE + DPP & 24.60 & 41.80 & 55.40 & 40.60 \\
\midrule
\textbf{HC}
& \textbf{24.40}
& \textbf{42.80}
& \textbf{66.76}
& \textbf{44.65} \\
\bottomrule
\end{tabular}
\caption{Component ablation of HC in the legal domain. PCE (10\%) directly selects a 10\% subset in Phase 1 using proxy-predicted scores, while PCE (20\%) trains on the full 20\% Phase-1 candidate pool without final pruning. Other variants replace either the candidate construction or the final pruning module.}
\label{tab:hc_component_ablation_law}
\vspace{-5mm}
\end{table}

\section{Sensitivity Analysis}
\label{app:parameter_sensitivity}

This section examines the effects of HC's scoring parameters and the final training ratio on downstream performance.

\subsection{Sensitivity to Scoring Parameters}
\label{app:scoring_parameter_sensitivity}

We evaluate how different combinations of $\alpha$, $\beta$, and $\gamma$ affect the selected data region and downstream fine-tuning performance. To isolate the effect of the scoring function itself, this experiment uses exact cartography scores computed from labeled data, rather than proxy-predicted scores. All variants maintain a constant final training budget and employ identical fine-tuning configurations.

We compare the following five parameter settings: (1) Equal C/V: $\alpha=0.50$, $\beta=0.50$, $\gamma=0.20$; (2) More Confidence: $\alpha=0.60$, $\beta=0.25$, $\gamma=0.15$; (3) More Variability: $\alpha=0.30$, $\beta=0.55$, $\gamma=0.15$; (4) Strong Penalty: $\alpha=0.45$, $\beta=0.35$, $\gamma=0.35$; (5) Weak Penalty: $\alpha=0.45$, $\beta=0.35$, $\gamma=0.05$.

Here, $\alpha$ controls the confidence term, $\beta$ controls the variability term, and $\gamma$ controls the extreme-sample penalty. These settings test whether HC should emphasize learnability, uncertainty, or robustness to extreme samples.

\begin{figure}[htbp]
\centering
\includegraphics[width=\columnwidth]{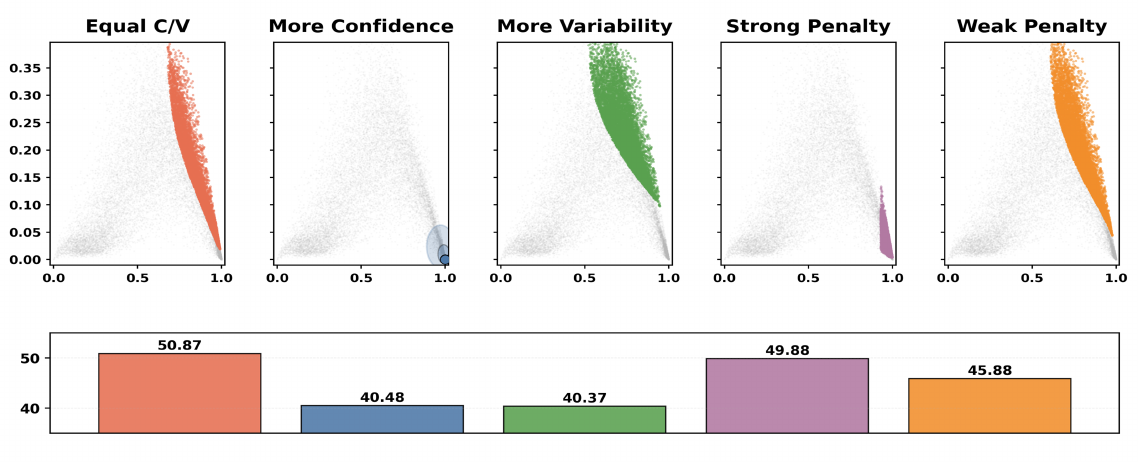}
\caption{Sensitivity analysis of HC scoring parameters on the medical domain. The top row shows the selected regions under different scoring parameters in the confidence-variability space. Gray points denote the full training pool, and colored points denote the selected subset. The bottom row reports the corresponding average downstream performance.}
\label{fig:parameter_sensitivity}
\end{figure}

Figure~\ref{fig:parameter_sensitivity} demonstrates that the Equal C/V setting achieves the best performance. The More Confidence setting concentrates selected samples in the high-confidence and near-zero-variability region, which primarily contains overly easy examples and yields suboptimal performance. Conversely, More Variability shifts the selection distribution toward highly unstable samples, similarly degrading performance. While Strong Penalty outperforms these two extreme settings by avoiding certain noisy or trivial regions, it still falls short of Equal C/V. Weak Penalty covers a broader region but proves less effective at filtering out extreme samples. Overall, these results indicate that high-value instruction-tuning data must strike a balance between learnability and informativeness while avoiding extreme regions. Consequently, HC adopts $\alpha=0.50$, $\beta=0.50$, and $\gamma=0.20$ as its default configuration.

\subsection{Effect of the Final Training Ratio}
\label{app:training_ratio_sensitivity}

\begin{figure}[htbp]
\centering
\includegraphics[width=0.95\linewidth]{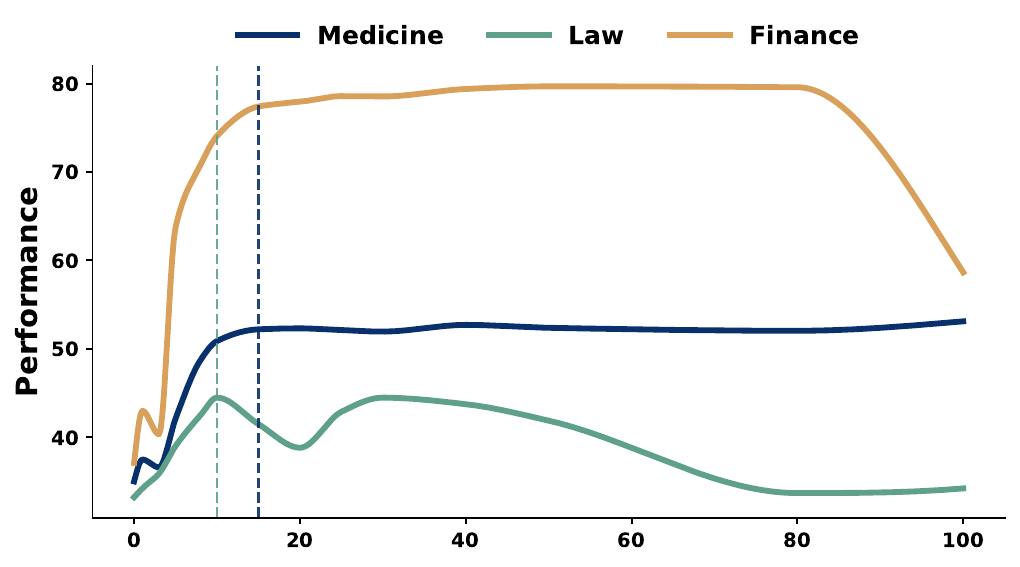}
\caption{Performance under different top-score training ratios in medicine, law, and finance. The x-axis shows the percentage of the full training pool retained for fine-tuning. Dashed vertical lines mark the final training ratios used in the main experiments: 15\% for medicine and finance, and 10\% for law.}
\label{fig:high_value_ratio}
\end{figure}

Figure~\ref{fig:high_value_ratio} illustrates downstream performance as a function of the proportion of top-scoring training samples retained. These ratios are measured relative to the full training pool in each domain. The dashed vertical lines mark the final training budgets used in our main experiments: 15\% for medicine and finance, and 10\% for law. Notably, increasing the retained data fraction does not consistently yield performance gains, highlighting a distinct trade-off between the final training-set size and downstream accuracy.

\end{document}